\documentclass[journal]{IEEEtran}
\usepackage{times}
\usepackage{epsfig}
\usepackage{graphicx}
\usepackage{amsmath}
\usepackage{amssymb}
\usepackage{subfig}
\usepackage{color}
\usepackage{hyperref}
\usepackage{booktabs}
\usepackage{authblk}

\newcommand{\etal}{\textit{et al}.}

\begin{document}
%
\title{Tracking in Aerial Hyperspectral Videos using Deep Kernelized Correlation Filters}
\author[*]{Burak Uzkent}
\author[*]{Aneesh Rangnekar}
\author[**]{Matthew J. Hoffman}
\affil[*]{Chester F. Carlson Center for Imaging Science, Rochester Institute of Technology}
\affil[**]{School of Mathematical Sciences, Rochester Institute of Technology}
\affil[ ]{\textit {\{bxu2522, apr2635, mjhsma\}@rit.edu}}
\maketitle

\begin{abstract}
Hyperspectral imaging holds enormous potential to improve the state-of-the-art in aerial vehicle tracking with low spatial and temporal resolutions. Recently, adaptive multi-modal hyperspectral sensors have attracted growing interest due to their ability to record extended data quickly from aerial platforms. In this study, we apply popular concepts from traditional object tracking, namely (1) Kernelized Correlation Filters (KCF) and (2) Deep Convolutional Neural Network (CNN) features to aerial tracking in hyperspectral domain. We propose the Deep Hyperspectral Kernelized Correlation Filter based tracker (DeepHKCF) to efficiently track aerial vehicles using an adaptive multi-modal hyperspectral sensor. We address low temporal resolution by designing a \textit{single KCF-in-multiple Regions-of-Interest (ROIs) approach} to cover a reasonably large area. To increase the speed of deep convolutional features extraction from multiple ROIs, we design an effective ROI mapping strategy. The proposed tracker also provides flexibility to couple with the more advanced correlation filter trackers. The DeepHKCF tracker performs exceptionally well with deep features set up in a synthetic hyperspectral video generated by the \textit{Digital Imaging and Remote Sensing Image Generation (DIRSIG)} software. Additionally, we generate a large, synthetic, single-channel dataset using DIRSIG to perform vehicle classification in the \textit{Wide Area Motion Imagery (WAMI)} platform. This way, the high-fidelity of the DIRSIG software is proved and a large scale aerial vehicle classification dataset is released to support studies on vehicle detection and tracking in the WAMI platform.\end{abstract}

\begin{IEEEkeywords}
vehicle tracking, hyperspectral sensing, deep features.
\end{IEEEkeywords}

%
\IEEEpeerreviewmaketitle
\section{Introduction}
\label{intro}
%
%
%
%
\IEEEPARstart{A}{erial} object tracking is a popular topic due to its large range of applications in security, traffic surveillance, autonomous driving, and UAV monitoring. Tracking from aerial platforms can be performed with a number of data modalities including, but not limited to, grayscale \cite{pelapur2012persistent}, thermal \cite{portmann2014people}, color \cite{danelljan2016eco} and most recently, hyperspectral imagery \cite{uzkent2015efficient,uzkent2016real,uzkent2017aerial}. Each modality has been exploited for a unique application and comes with its own set of advantages and disadvantages. For the scope of this paper, we will focus on two specific types of sensor modalities: (1) Wide Area Motion Imagery (WAMI), and (2) \textit{Adaptive} Hyperspectral Imagery. 

The WAMI platform can scan up to a $5$ km $\times$ $5$ km area at 2 frames per second (fps), with vehicles occupying roughly $100$ - $200$ pixels. One can perform persistent tracking utilizing this large field of view image. The low spatial resolution of WAMI has good performance in vehicle tracking under certain circumstances but is not super helpful when it comes to handling background clutter, occlusions, and low contrast objects. To counter this, multiple appearance-based features like textures, color histograms, and histogram of gradients, and motion cues are used in combination, leading to multiple heat maps for the search area. This is not feasible in real-time tracking as the methods can become computationally expensive and hence there exists a need to balance between designing complex models and facilitating real-time implementations. The datasets recorded by WAMI are: (1) WPAFB 2009 \cite{WPAFB} and (2) CLIF \cite{CLIF}, where each one includes a single video with less than 100 frames. This is detrimental towards training standalone machine learning and deep learning based architectures since the amount of available data is quite low for training purposes. Since data collection from an aerial platform is lengthy and costly, it is possible to utilize deep learning architectures as feature encoders rather than an end-to-end tracking system. However, WAMI is not a practical platform since it provides single-channel imagery and the deep learning architectures are trained on ImageNet \cite{russakovsky2015imagenet} which has RGB images.

\begin{figure*}[t]
\centering
\includegraphics[width=\textwidth]{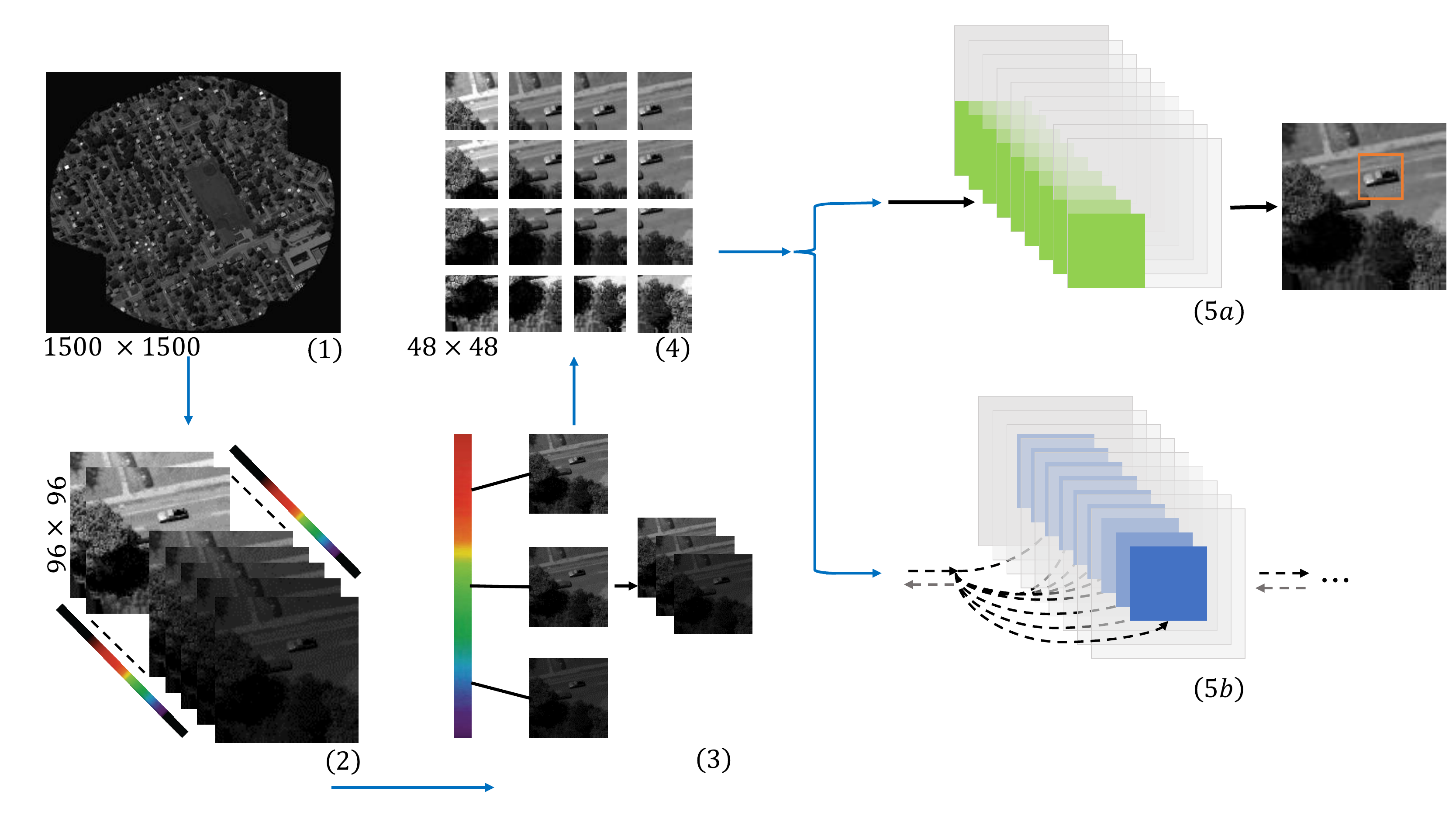}\\
\caption{The proposed Kernelized Correlation Filter driven tracker inspired by RITMOS. In Step 1, registration is performed by using single-channel large field of view (FOV) imagery. The small FOV hyperspectral image (61 bands in 400-1000 nm wavelength range) is then sampled in Step 2 to detect the target. The region of interest (ROI) is then divided into subregion of interests (48 $\times$ 48 px) to run kernelized correlation filter (KCF) on each of them in Step 4 and detect the target as in Step 5a. Finally, the KCF model is updated in Step 5b before passing to the next frame. The highlighted tracker represents the FastDeepHKCF that forward-propagates the ROI to the layer of interest in the CNN and projects the subregions to the feature maps. With RITMOS, it costs about 0.1s. to generate a single channel image in step 1 and another 0.1s for a small field of view hyperspectral image in step 2.}
\label{fig:tracking_overview}
\end{figure*}

The unique challenges posed by aerial platforms can be better addressed by smarter multi-modal data acquisition. In this direction, the \textit{Rochester Institute of Technology Multi-object Spectrometer} (RITMOS) concept is utilized by a number of trackers \cite{uzkent2013feature,uzkent2015feature,uzkent2016real,uzkent2017aerial} as an example that can collect a small, targeted amount of hyperspectral data. The RITMOS captures data in two different modalities : (1) a full frame single channel image, and (2) limited hyperspectral data from the desired pixel locations. It can acquire a full-frame single channel image in about $0.1$ sec and scan a row of pixels hyperspectrally in $1$ ms. Such an adaptive and multi-modal data concept provides more freedom to address aerial tracking challenges. Driven by this freedom and specifications, we design a \textit{discriminative} tracker to operate on this platform as shown in Fig.~\ref{fig:tracking_overview}. We refer the readers to \cite{meyer2004ritmos, uzkent2016real} for more information on the workings of RITMOS.

\begin{figure}[h]
\hfill
\subfloat[$130$th frame (\textit{with trees})]{\includegraphics[width=0.24\textwidth]{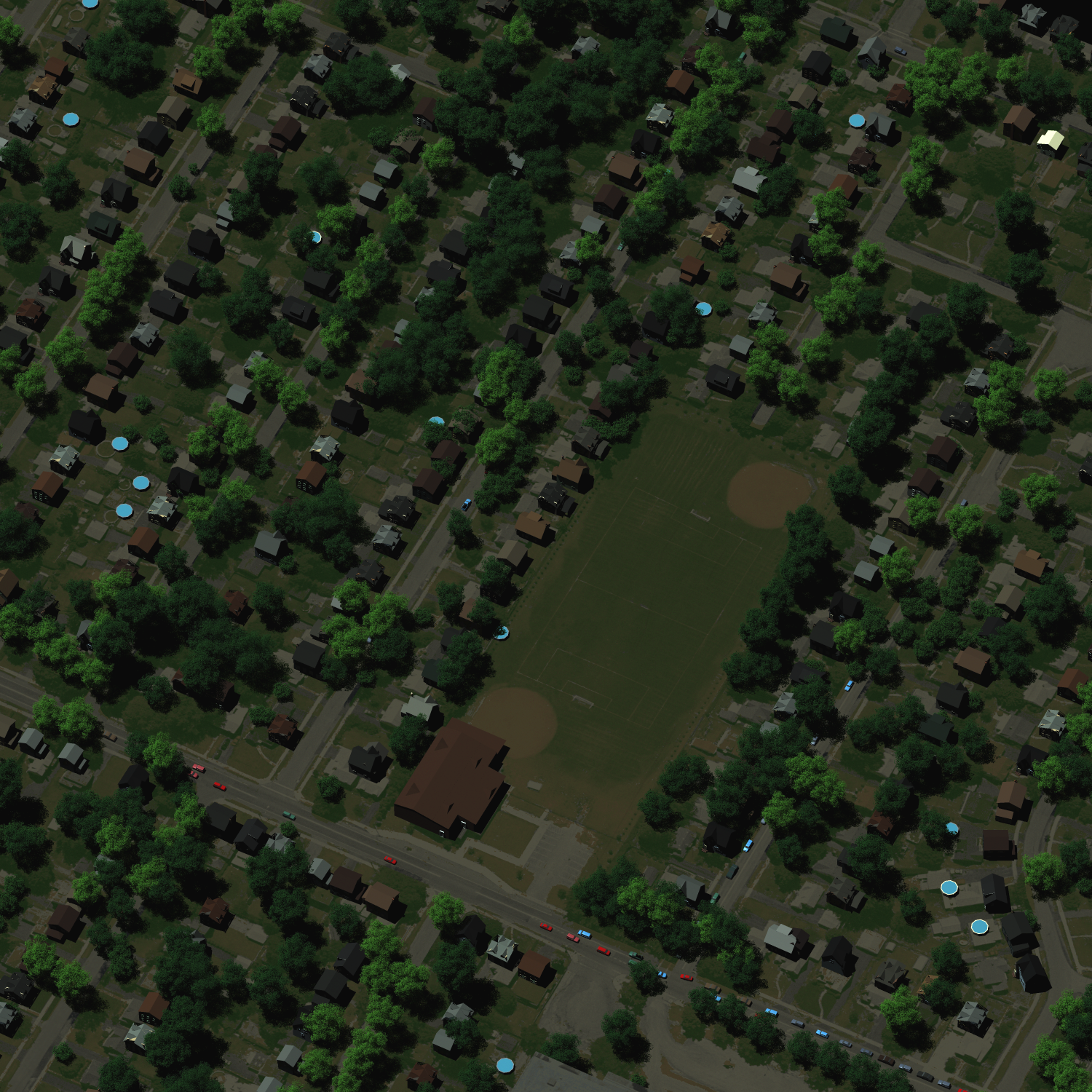}}
\hfill
\subfloat[$130$th frame (\textit{without trees})]{\includegraphics[width=0.24\textwidth]{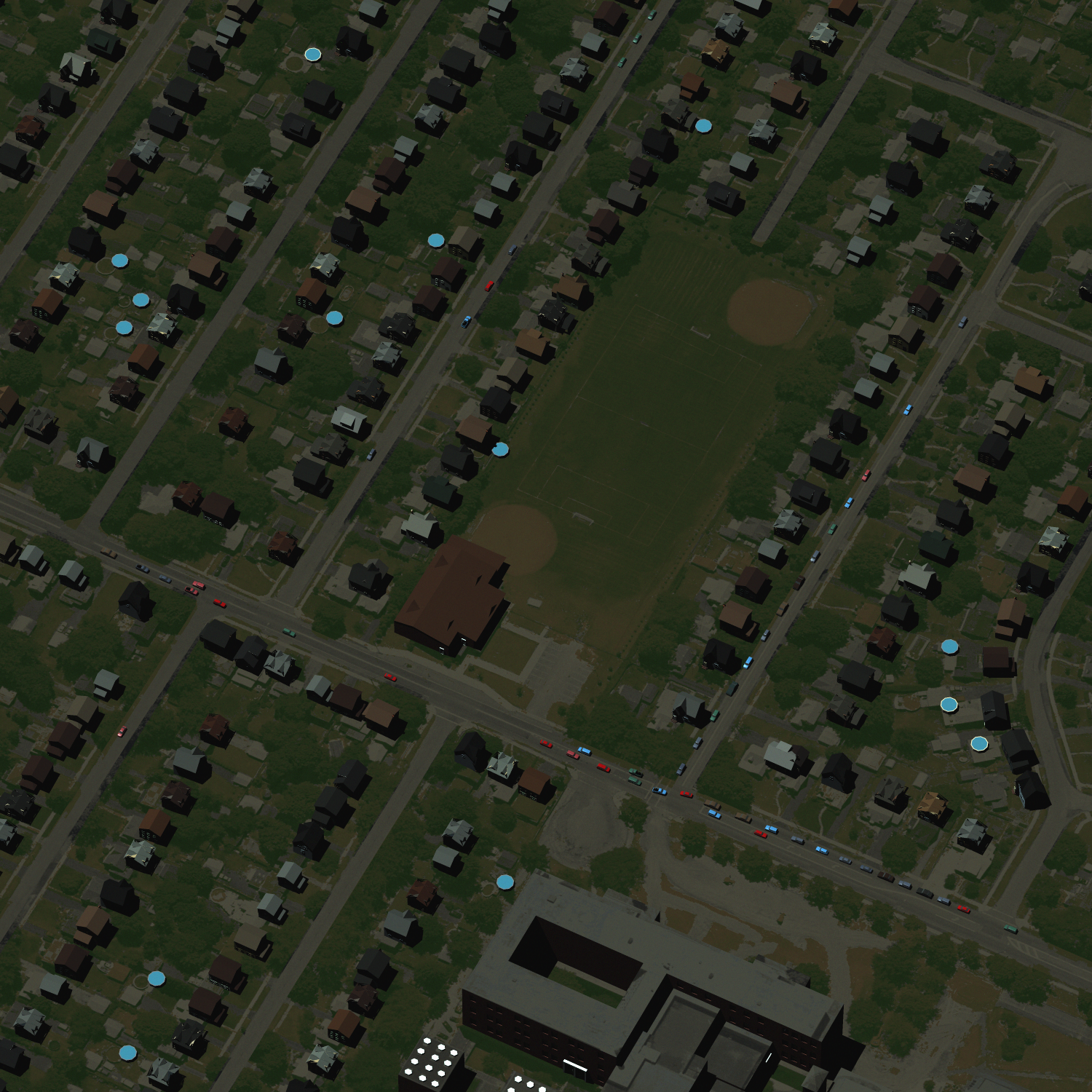}}
\hfill
\caption{Two frames from the synthetic scenarios generated by DIRSIG. The scene comes from the Mega-Scene I area available in DIRSIG. The Mega-Scene I area represents part of Rochester, NY.}
\label{fig:dirsig_scenarios}
\end{figure}

\begin{flushleft}
\textbf{Synthetic Imagery concept}
\end{flushleft}
The \textit{Digital Imaging and Remote Sensing (DIRSIG)} software has been used before to generate spectral scenarios for varied applications that use conventional computer vision techniques and deep learning based models \cite{uzkent2016integrating,uzkent2017aerial,han2017efficient,han2017overview}. Since flying spectral sensors on an aerial platform is still an ongoing area of development due to the high costs involved, we evaluate our tracker on synthetic scenarios generated using DIRSIG by \cite{uzkent2015spectral,uzkent2016real_2,uzkent2016real}. In particular, we focus on two scenarios: (1) with trees and (2) without trees. We track \textit{43} vehicles in both scenarios as shown in Fig.~\ref{fig:dirsig_scenarios}. In addition, we generate a synthetic single-channel aerial dataset for training a CNN and use it to perform vehicle classification on the real WAMI platform, similar to \cite{han2017efficient} (Sect.~\ref{sect:vehicle_classification}).

\begin{flushleft}
\textbf{Motivations}
\end{flushleft}
The rich sensory information from hyperspectral imagery has been utilized by \textit{generative} trackers \cite{uzkent2016real,uzkent2017aerial}. The \textit{discriminative} and \textit{deep learning} driven trackers on the other hand have recently improved the traditional object tracking dramatically. The main challenges behind the application of discriminative and deep learning trackers in aerial hyperspectral images are:

\begin{itemize}
\item Well-established discriminative algorithms such as Efficient Convolutional Operators (ECO) \cite{danelljan2016eco}, Kernelized Correlation filters (KCF) \cite{henriques2015high}, Struck \cite{hare2016struck}, and Tracking-learning-detection (TLD) \cite{kalal2012tracking} are mostly associated with close-angle color image single/multi object tracking at high video frame rates and thus, consider a small region of interest (ROI). 
\item The cost of collecting data from aerial platforms makes it hard to find/collect large samples of aerial data, leaving the community to split a single video into training and validation sets \cite{lalonde2017fully, yi2016vehicle}. This leads to very optimistic results during off-line tracking due to minimal variance in the training dataset, which results poor performance during online tracking. 
\end{itemize}

\section{Related Work}
\label{Related_Work}
Tracking-by-detection algorithms exploited low-level features such as Histogram of Oriented Gradients and Color-naming features \cite{dalal2005histograms,li2014scale,felzenszwalb2010object} 
to perform discriminative tracking until the emergence of deep CNN architectures in the computer vision field. The first correlation filter tracker - the Minimum Output Sum of Squared Error (MOSSE) filter \cite{bolme2010visual}, used a grayscale channel to learn the classifier vector. Following the MOSSE tracker, a correlation filter accommodating multi-channel features was proposed to boost tracking \cite{galoogahi2013multi}. Later, the Scale Adaptive with Multiple Features (SAMF) tracker \cite{li2014scale} was proposed to concatenate multi-channel HoG  and color-naming features. Finally, a kernelized version of correlation filter (KCF) using  multi-channel features was proposed to further improve tracking without drastically increasing the computational complexity \cite{henriques2012exploiting,henriques2015high}. 

The first studies utilizing CNN architectures in object tracking focused on employing the features learned in architectures such as AlexNet \cite{krizhevsky2012imagenet}, VGGNet \cite{simonyan2014very} trained on the ImageNet dataset \cite{russakovsky2015imagenet}. \cite{danelljan2015convolutional} extracted low-level features from VGGNet to learn a more discriminative correlation filter. Specifically, they encoded objects with the activations of the first several convolutional layers from VGGNet. This setting provided them with a $64$ $\times$ $64$ $\times$ $96$ dimensional low-level feature set that can be interpreted as a more advanced version of HoG features. They reported slight improvement in the Visual Object Tracking Challenge 2015 (VOT2015) object tracking challenge with deep CNN features over the HoG features. Going deeper is a major key to achieving the state-of-the-art in most computer vision challenges, however, the nature of deep CNN architectures prohibits us from applying high-level features in tracking-by-detection algorithms. This is mainly due to increasing \textit{translation invariance} in deeper layers resulting from spatial pooling operations. 

The object tracking community later migrated to training architectures to perform object tracking in an end-to-end framework \cite{held2016learning, bertinetto2016fully, leal2016learning}. In this direction, \textit{Siamese Networks} have gained the reputation of the most efficient and effective architecture in tracking. Two branches consisting of the same architecture layers are used in a typical Siamese Network. The bottom branch is provided the ground truth of an object of interest in an ROI, whereas the top branch is assigned the task of estimating the position of the object given the new ROI. Late fusion of the branches is performed and the new position is regressed. The Siamese Networks have surpassed all the other deep tracking-by-detection algorithms in the VOT2015 challenge. Due to the scarcity of annotated datasets for aerial tracking, it is difficult to develop an end-to-end deep learning tracker for aerial platforms. 

There is scarcity of annotated datasets for aerial tracking in which deep learning or traditional trackers can be trained and evaluated. UAV123 \cite{mueller2016uav123}, recently released by Meuller \etal, has a ground sampling distance (GSD) that is significantly lower than the high-altitude aerial platforms - thus resulting in objects occupying more than $500$ - $1000$ pixels. The dataset has sequences at 30 fps, drastically higher than standard WAMI and spectral sequences, which are generally in the 1.42 fps - 2 fps range. Flying RITMOS on an aerial platform is still an ongoing area of development and due to lack of any other real dataset in this area, we use synthetically generated \textit{Rochester Institute of Technology Multi-object Spectrometer} (RITMOS)-like data to evaluate the performance of our proposed tracker. This way, we prevent probable overfitting that would have been caused due to training and testing on the same dataset by using deep learning models as feature encoders in our tracker.

\begin{flushleft}
\textbf{Contributions}
\end{flushleft}
This study addresses the unique challenges posed by the application of discriminative trackers to aerial platforms. A novel method that employs a discriminative tracker is proposed to tackle low temporal (around 1.42 fps) and spatial resolutions (0.3 m). Primarily, we design a method to enlarge the area considered by the tracker to handle the low temporal resolution. Given the rich hyperspectral imagery, we utilize pre-trained deep convolutional networks as feature encoders to boost tracking performance. To accommodate deep features in a near real-time tracking system, we design a \textit{region-of-interest (ROI) mapping} strategy that only forward passes the large ROI and projects the individual ROIs to the large ROI feature maps (Fig. \ref{fig:tracking_overview}). Finally, the proposed tracker is evaluated on a synthetic hyperspectral video generated by the Digital Imaging and Remote Sensing (DIRSIG) software \cite{ientilucci2003advances}. To prove the high-fidelity of this video, a large single-channel aerial dataset is synthesized using DIRSIG and a deep learning framework is trained on it to classify images from the real dataset (WAMI). We refer the readers to following link to access our synthetic vehicle classification dataset (\href{https://buzkent86.github.io/datasets/}{\color{blue}https://buzkent86.github.io/datasets/}).

To the best of our knowledge, this is the first time an adaptive hyperspectral sensor-inspired discriminative tracker (DeepHKCF) has been proposed to perform robust single target tracking in spectral aerial imagery that can be generalized to the WAMI platform. 

\section{Proposed Tracker}
\label{proposed_algorithm}
As discussed in the previous section, the tracking platform has low frame rate making the global camera motion removal step necessary to perform consistent tracking. In this direction, we register the input frame to the canonical frame where the tracking is initialized using standard computer vision techniques. First, keypoints in the images are extracted using the Scale Invariant Feature Transform (SIFT) \cite{lowe2004distinctive} and described with gradient orientation histograms. In the next step, the homography matrix between two images is estimated with the Random Sample Consensus (RANSAC) \cite{fischler1987random} algorithm. Finally, the input image is warped to the canonical image using the accumulated homography matrix over time.

The core of the proposed tracker is built upon the work of Henriques \etal \cite{henriques2012exploiting,henriques2015high} with Kernelized Correlation Filters (KCF). The KCF has emerged as a high accuracy tracker that can operate at hundreds of frame rate under specific conditions. Its computational efficiency is derived from the correlation filter framework that represents training examples using a circulant matrix. The fact that a circulant matrix can be diagonalized by Discrete Fourier Transform (DFT) is the key to reducing the complexity of any tracking method based on the correlation filter. The off-diagonal elements become zero whereas the diagonal elements represent the eigenvalues of the circulant matrix. The KCF applies a kernel to transform the feature channels to a more discriminative domain. 

Essentially, the KCF solves the problem in the form of ridge regression:
\begin{equation}
E(w) = \frac{1}{2}||y-\sum_{c=1}^{C}(w_{c}*x_{c})||^{2} + \frac{\lambda}{2}\sum_{c=1}^{C}||w_{c}||^{2}
\label{eq:Closedform_RidgeReg}
\end{equation}
where $y$ represents the desired continuous response, $w$ represents the correlation filter 
and $x_{c}$ represents template for the given channel. The parameter $C$ enables one to
integrate features in multiple channel space: an earlier version based on this formulation employed grayscale feature ($C = 1$) to learn the solution vector $w$. Later, multi-channel features such as Color, HoG and a concatenation of them showed improved accuracy \cite{henriques2015high,galoogahi2013multi,tang2015multi,ma2015long,bibi2015multi}. To reduce the complexity of the closed-form solution for Eqn. \ref{eq:Closedform_RidgeReg}, an element-wise multiplication in the frequency domain was proposed for $\hat{w}$ \cite{bolme2010visual}:
\begin{equation}
\hat{w}=\frac{\hat{x}^{*}\odot\hat{y}}{\hat{x}^{*}\odot\hat{x}+\lambda},
\label{eq:DiagonalizedPrimalSolution}
\end{equation}
where $\hat{\,}$ and $*$ denote the parameter in Fourier domain and conjugate of a complex number whereas $\odot$ and $\lambda$ are the element-wise multiplication, and a regularization term to prevent divisions by zero.

The solution to the kernelized version of ridge regression is given
by \cite{rifkin2003regularized} as follows:
\begin{equation}
\alpha=\left(K+\lambda I\right)^{-1}y
\end{equation}
where $K$ is the kernel matrix and $\alpha$ is the vector of coefficients $\alpha_{i}$, that represent the solution in the kernel-transformed dual space.
The diagonalized Fourier domain dual form solution (non-linear version) is then expressed as 
\begin{equation}
\hat{\alpha} = \hat{y}(\hat{k}^{xx}+\lambda)^{-1},
\label{eq:FourierDualDomainSolution}
\end{equation}
where ${k}^{xx}$is the first row of the kernel matrix $K$ and is the kernel's autocorrelation.


For multiple channel cases, we obtain $\hat{k}^{{xx}^{'}}$, which represents the first row of the kernel matrix $K$ in the frequency domain, also known as \textit{gram matrix}. It can be formulated as:
\begin{equation}
k^{xx^{'}} = exp(-\dfrac{1}{\alpha^{2}}(||x||^{2}+||x^{'}||^{2}-2F^{-1}(\sum^{c = C}_{c = 1}\hat{x}_{c}^{*}\odot \hat{x}_{c}^{'}))).
\label{eq:GaussianCorrelationSingleChannel}
\end{equation}
where $x$ concatenates the individual vectors for $C$ channels: $x=\left[{x}_{1},\ldots,{x}_{C}\right]$. In training step, the arbitrary vector $x^{'}$ is replaced by $x$, and in test step, it is replaced by $z$.
 
To detect the object of interest, we typically wish to evaluate the regression function $f({z})$ on several locations in the image, i.e. several candidate patches, which can be modeled by cyclic shifts.
\begin{equation}
f({z})={w}^{T}{z}=\sum_{i=1}^{n}\alpha_{i}\kappa({z},{x}_{i})
\label{eq:regzequation}
\end{equation}




Since ${f}({z})$ is a vector containing the output for all cyclic shifts of ${z}$, we can diagonalize it to obtain a more efficient computation in the Fourier domain:
\begin{equation}
\hat{{f}}({z})=\hat{{k}}^{{xz}}\odot\hat{{\alpha}}\label{eq:fast-detection},
\end{equation}
where ${k}^{{xz}}$ is the kernel correlation of ${x}$ and ${z}$.

Eqn. \ref{eq:fast-detection} then translates into the following equation in time domain: 
\begin{equation}
r(z) = F^{-1}(\hat{k}^{xz} \odot \hat{\alpha})
\label{eq:rzcorr}
\end{equation}
where $r$ denotes the correlation response at all cyclic shifts of the
first row of the kernel matrix. 

The temporal information can be further integrated into the tracker by updating the
filter and target template at every frame as follows:
\begin{align}
\hat{\alpha}_{t} = (1-\beta)\hat{\alpha}_{t-1} + \beta \hat{\alpha}_{t}, \\
\hat{x}_{t} = (1-\beta)\hat{x}_{t-1} + \beta\hat{x}_{t}
\end{align}
where $\beta$ is the learning rate. This correlation filter framework only estimates the translation of the object whereas the scale of the object can be updated by running a correlation filter on different size ROIs with same centroids \cite{li2014scale}. By correlating a filter with different ROIs, we can get multiple response maps and choose the one with highest confidence to estimate the new scale of the target. In this study, we do not estimate the scale of the target as the scenarios are captured from a fixed altitude platform.

\subsection{Single KCF-Multiple ROIs Approach}
Discriminative trackers like KCF learn to function in an online manner by collecting positive and negative samples and then detecting the target of interest in a ROI to update the classifier. The standard form KCF requires small ROIs as the appearance-based features deteriorate with larger background context. Unfortunately, these features are hard to collect from aerial imaging platforms due to their low spatial resolution. Moreover, there are two other limitations: (1) Increasing the context size leads to background dominated features resulting in confusion between different objects and (2) The platform we consider has lower temporal resolution ($1.4$ fps) leading to large displacement of objects in successive frames. Adding the platform motion into this picture makes the application of vanilla-form KCF in aerial platforms extremely difficult.
\begin{figure}[!h]
\centering
\includegraphics[width=0.5\textwidth]{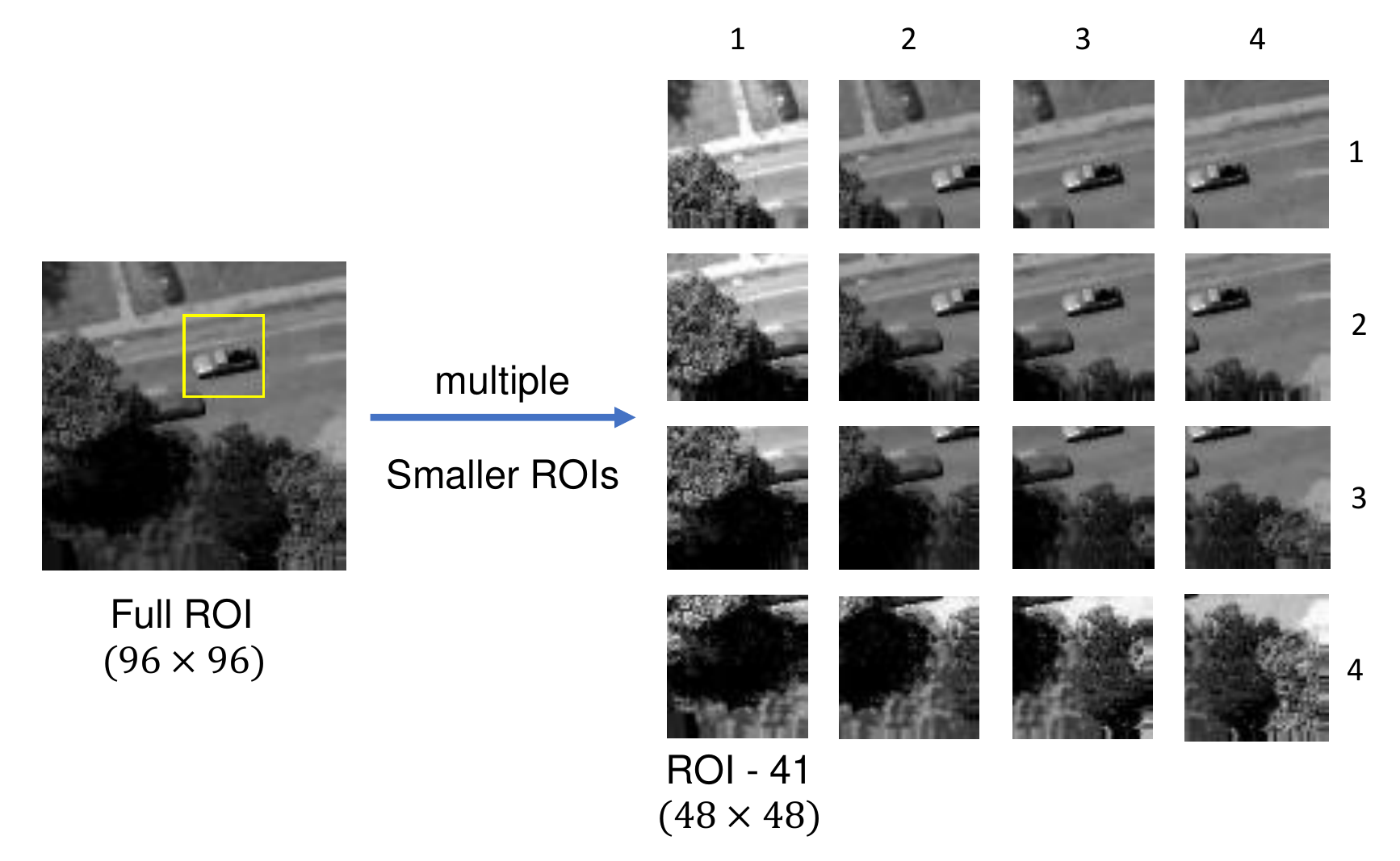}\\
\caption{The proposed single KCF - multiple ROI approach to enlarge the ROI to overcome large displacement of vehicles in low resolution data with 64\% window overlap (See sect.\ref{overlap_ratio_exps}). The vehicle of interest is shown in yellow rectangle.} 
\label{KCF_Grid_Approach}
\end{figure}
To handle these challenges, we propose a single KCF-in-multiple ROIs approach (Fig.~\ref{KCF_Grid_Approach}). Our approach applies the same KCF to different ROIs overlapping each other to minimize the likelihood of target loss. It is essential to have reasonable overlap between the ROIs (Sect.~\ref{overlap_ratio_exps}) as we filter each ROI with a Hanning window to avoid distortion at boundaries due to \textit{FFT} operation. This approach can be formulated by modifying Eqn. \ref{eq:rzcorr} as:
\begin{equation}
r(z_{ij}) = F^{-1}(\hat{k}^{xz_{ij}} \odot \hat{\alpha})
\label{eq:response_map}
\end{equation}
where $i$ and $j$ represent the indexes for different ROIs. A simple way to estimate the new position of the target in this framework would be
using the peak-to-side-lobe ratio (PSR) values in ROIs and finding the position of the pixel with maximum confidence in all ROIs as: 
\begin{equation}
r(z_{final}) = \mathbf{\underset{i,j\in \sqrt{m}}{\textit{argmax}}(PSR(r(z_{ij})))}
\label{eq:hard_decision}
\end{equation}
where $m$ represents the number of ROIs in full ROI. The \textit{PSR}, on the other hand, denotes the margin between the peak value in the response map and the mean of the sidelobe corresponding to the area excluding the $11$x$11$ pixels around the peak. The result is normalized by the standard deviation of the sidelobe as follows.
\begin{equation}
PSR(r(z_{ij})) = \frac{max(r(z_{final})) - \mu_{sidelobe}}{\sigma_{sidelobe}} 
\end{equation}

This position estimation approach can be softened by considering all the ROIs with PSR values larger than a pre-determined threshold, $T$. In this case, the Eqn.~\ref{eq:hard_decision} can be reorganized as follows.
\begin{align}
&r(z_{final}) = \sum_{i}^{\sqrt{m}} \sum_{j}^{\sqrt{m}} \beta_{ij} * r(z_{ij}), \label{eq:final_response_map}\\
&\beta_{ij} = \left\{\begin{array}{l l}
    0, \quad \textrm{if}\:PSR(r(z_{ij}))>T \\
    PSR(r(z_{ij})), \quad \textrm{otherwise}.
  \end{array}
\right.
\end{align}

By softening our decision, we perform low-pass filtering and avoid jumps to other objects that has a high PSR value in only one ROI.

As mentioned earlier, the single KCF-multiple ROIs approach can better handle the low temporal resolution than the traditional KCF. On the other hand, it increases the complexity linearly from $O(n log(n))$ to $O(m * n log(n))$, where $m$ represents the number of ROIs in the full ROI. The low temporal frame rate of the scenario helps us accommodate this approach in the DeepHKCF tracker. It is possible to further increase the frame rate by running the KCF on the multiple ROIs in parallel as the ROI operations are independent.

\subsection{Traditional Low-level Features}
In this study, we follow the KCF tracker and concatenate multiple features as in the SAMF \cite{li2014scale} tracker. More specifically, we concatenate the Felzenszwalb's HoG (fHoG) \cite{felzenszwalb2010object} feature channels and pure hyperspectral channels and apply the Gaussian kernel operation to learn a more discriminative model as follows:
\begin{align}
&k^{xz} = exp(-\dfrac{1}{\alpha^{2}}(||x||^{2}+||z||^{2}-  \\ \nonumber 
&2F^{-1}(\sum^{C_{fHoG}}_{c_{fHoG}=1}\hat{x}_{c_{fHoG}}^{*}\odot \hat{z}_{c_{fHoG}}+ \sum^{C_{HSI}}_{c_{HSI}=1}\hat{x}_{c_{HSI}}^{*}\odot \hat{z}_{c_{HSI}})
\label{eq:GaussianCorrelationSingleChannel}
\end{align}
where $c_{HSI}$ and $c_{fHoG}$ represent the hyperspectral and fHoG feature channels. The number of hyperspectral and fHoG feature channels in this study are 61 and 31 respectively. Additionally, in the results section (Sect.~\ref{section:Tracking Experiments}), we experiment with fHoG features and hyperspectral features alone to observe how well they perform individually.

\subsection{Deep Convolutional Features}
In this study, we follow an approach similar to \cite{danelljan2015convolutional} to learn a discriminative model for the KCF. Low spatial resolution scenario enables us to pursue a slightly higher level of abstraction of objects. In particular, we apply the activations of the fifth convolutional layer learned in VGGNet \cite{simonyan2014very} trained over ImageNet \cite{russakovsky2015imagenet}. Additionally, we experiment with different levels of object abstractions in the experiments section (Sect.~\ref{section:Tracking Experiments}).

DIRSIG imagery provides us with a full-frame grayscale image as well as a narrow field of view hyperspectral image at 1.42 fps. Unlike other aerial platforms, it provides hyperspectral data in the visible wavelength range, enabling the use of deep CNN architectures trained on ImageNet consisting of RGB images. One can pick the central \textit{red}, \textit{green} and \textit{blue} channels and forward-pass them through the layers of interest. Another approach could be computing the average of \textit{red}, \textit{green} and \textit{blue} channels in their respective range to come up with the representative red, green and blue channel images to feed the CNN. Our experiments favor the first approach as the latter approach introduces undesired noise due to the averaging operation.

\begin{figure}[t]
\centering
\subfloat[ROI mapping for training in FastDeepHKCF tracker.]{\includegraphics[width=0.5\textwidth]{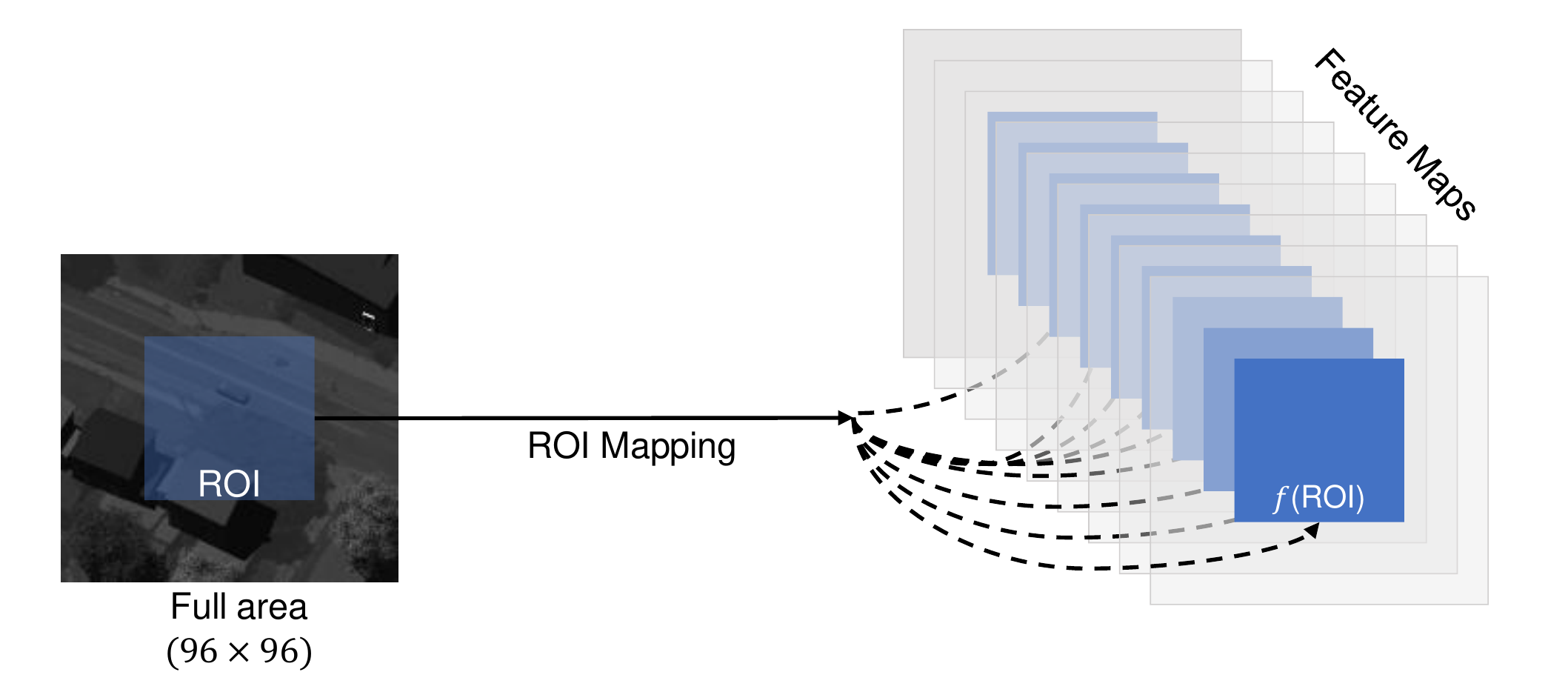}\label{fig:roi_mapping_train}}
\hfill
\subfloat[ROI mapping for detection in FastDeepHKCF tracker.]{\includegraphics[width=0.5\textwidth]{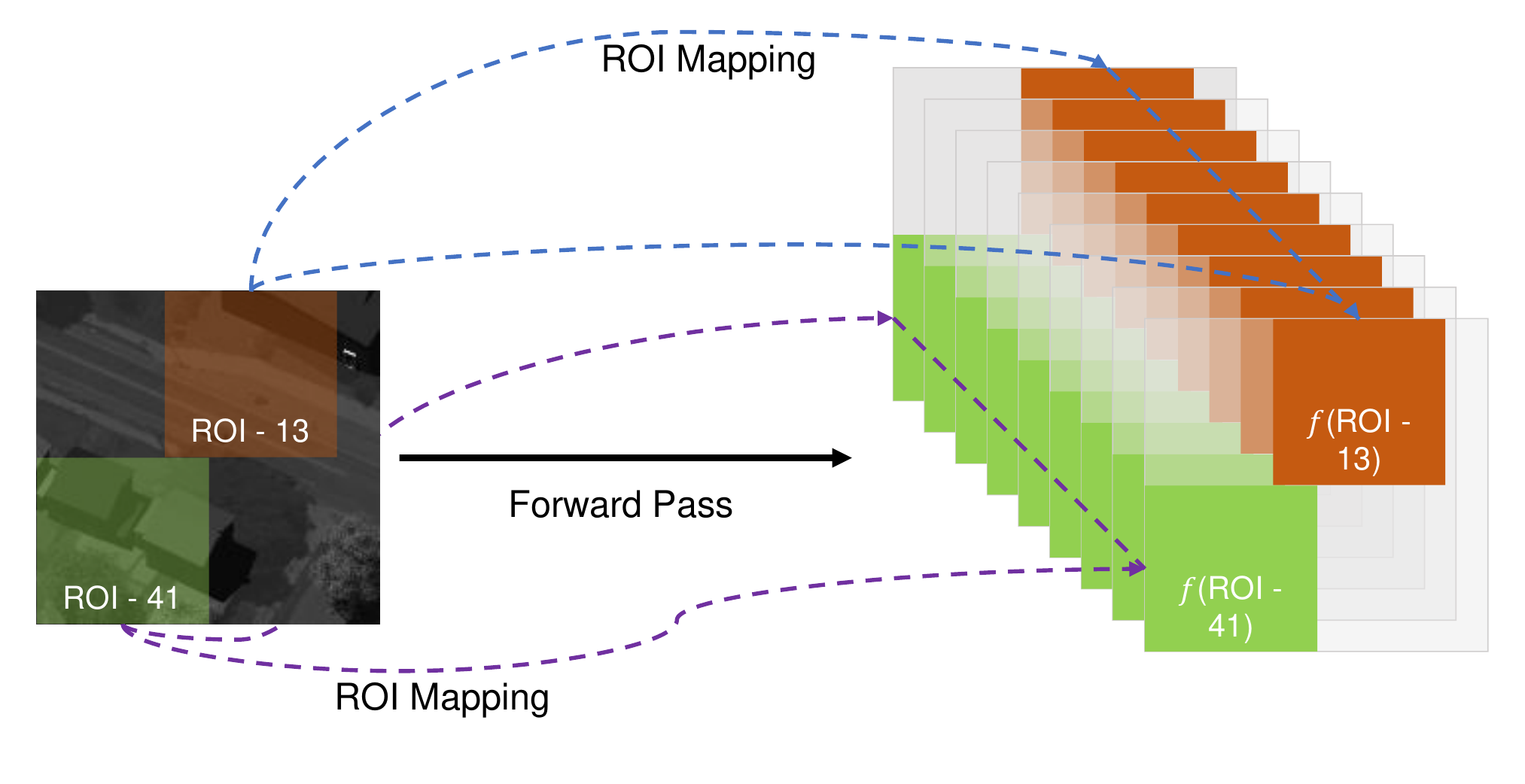}\label{fig:roi_mapping_detection}}
\hfill
\caption{Proposed ROI mapping strategy to avoid individual ROIs forward-passing through the convolutional network.}
\label{fig:roi_mapping}
\end{figure}

\begin{flushleft}
\textbf{Fast Convolutional Features with ROI Mapping}
\end{flushleft}
The single KCF-multiple ROIs approach treats each ROI independently to compute the filter response. This requires forward-passing individual ROIs through the CNN architecture. Such an inefficient approach leads to a slower tracker. To increase the run-time performance and perform near real-time tracking at the platform frame-rate, we use the ROI mapping strategy commonly used in convolutional object detectors such as Fast R-CNN \cite{girshick2015fast}, Faster R-CNN \cite{ren2015faster}, and R-FCN \cite{dai2016r}. This way, we only forward-pass the full ROI and project the individual ROIs to the feature maps extracted from the full ROI as shown in Fig.~\ref{fig:roi_mapping}.

With the ROI mapping method, Eqns.~\ref{eq:response_map} and~\ref{eq:final_response_map} can be replaced with the following formulations to perform detection in the FastDeepHKCF tracker.
\begin{align}
r(f(dROI,droi_{ij},z)) &= F^{-1}(\hat{k}^{xf(dROI,droi_{ij},z)} \odot \hat{\alpha}), \\
r(z_{final}) = \sum_{i}^{\sqrt{m}} \sum_{j}^{\sqrt{m}}& \beta_{ij} * r(f(dROI,droi_{ij},z))
\end{align}
where $dROI$ represents the full detection ROI used to get the convolutional features $z$. The individual detection ROIs, $droi_{ij}$, are then projected to the feature map, $z$, through the projection function $f$. 

Once we estimate the translation of the target, the filter is updated using the Fourier domain solution as in Eqn.~\ref{eq:FourierDualDomainSolution}. First, the $96$ $\times$ $96$ px neighborhood around the target is considered and forward-passed through the convolutional network as shown in Fig.~\ref{fig:roi_mapping_train}. To match the detection ROI size (Fig.~\ref{fig:roi_mapping_detection}), we then project the central $48$ $\times$ $48$ px area to the feature maps and reformulate the solution as:
\begin{equation}
\hat{\alpha} = \hat{y}(\hat{k}^{f(tROI,troi,x)f(tROI,troi,x)}+\lambda)^{-1}
\label{eq:FourierDualDomainSolution_fastdeephkcf}
\end{equation}
where $tROI$ and $troi$ represent the full training ROI and actual training ROI mapped to feature maps of the $tROI$ using the function $f$. On the other hand, we can avoid forward-passing the training ROI if the actual training ROI, $troi$, is a subset of the detection ROI, $dROI$. In this case, the Fourier domain solution can be reformulated as
\begin{equation}
\hat{\alpha} = \hat{y}(\hat{k}^{f(dROI,troi,z)f(dROI,troi,z)}+\lambda)^{-1}.
\label{eq:FourierDualDomainSolution_fastdeephkcf_wofp}
\end{equation}


\section{Vehicle Classification in WAMI Platform by using a Synthetic Dataset}
\label{sect:vehicle_classification}
As discussed in Section ~\ref{intro}, detecting cars with high accuracy is a major problem of tracking algorithms utilizing the \textit{WAMI} platform. This is due to two major reasons : (1) the lack of a large dataset captured from the WAMI platform and (2) the lack of color channels prevents smooth transfer learning from the networks trained on the ImageNet. 

In this study, we build a synthetic single-channel vehicle classification dataset using \textit{DIRSIG} and fine-tune a CNN to perform vehicle classification on the real platform (WAMI). To build this dataset, we generate full-frame hyperspectral images captured from the \textit{Mega-Scene I} scene available in DIRSIG \cite{ientilucci2003advances} with different settings. The simulation setting is designed as a function of \textit{time}, and hence the brightness in the scene varies as a function of sunlight which can then lead to a more general dataset. In particular, nine simulations from different months in a year are generated to find representative samples of varying conditions. We keep the other parameters similar to the simulation used to generate the RITMOS-like scenario. Overall, \emph{nine} different simulations are generated from the same scene with the same vehicular traffic and platform motion to the tracking video. 

\subsection{Temporal Data Augmentation}
It is possible to generate just one frame per simulation. However, to increase the number of simulations, we add more temporal-variance and change the initial platform location. Changing the platform location in a large number of simulations can be a tedious task, and to avoid that, we perform temporal data augmentation by generating low frame rate videos on a moving platform. More specifically, the frame rate for each simulation is set to $0.2$ fps, resulting in $20$ images per simulation. This way, we can capture cars from different angles with different backgrounds. 

\subsection{Hyperspectral Data Augmentation}
\begin{figure*}[!h]
\centering
\includegraphics[width=0.85\textwidth]{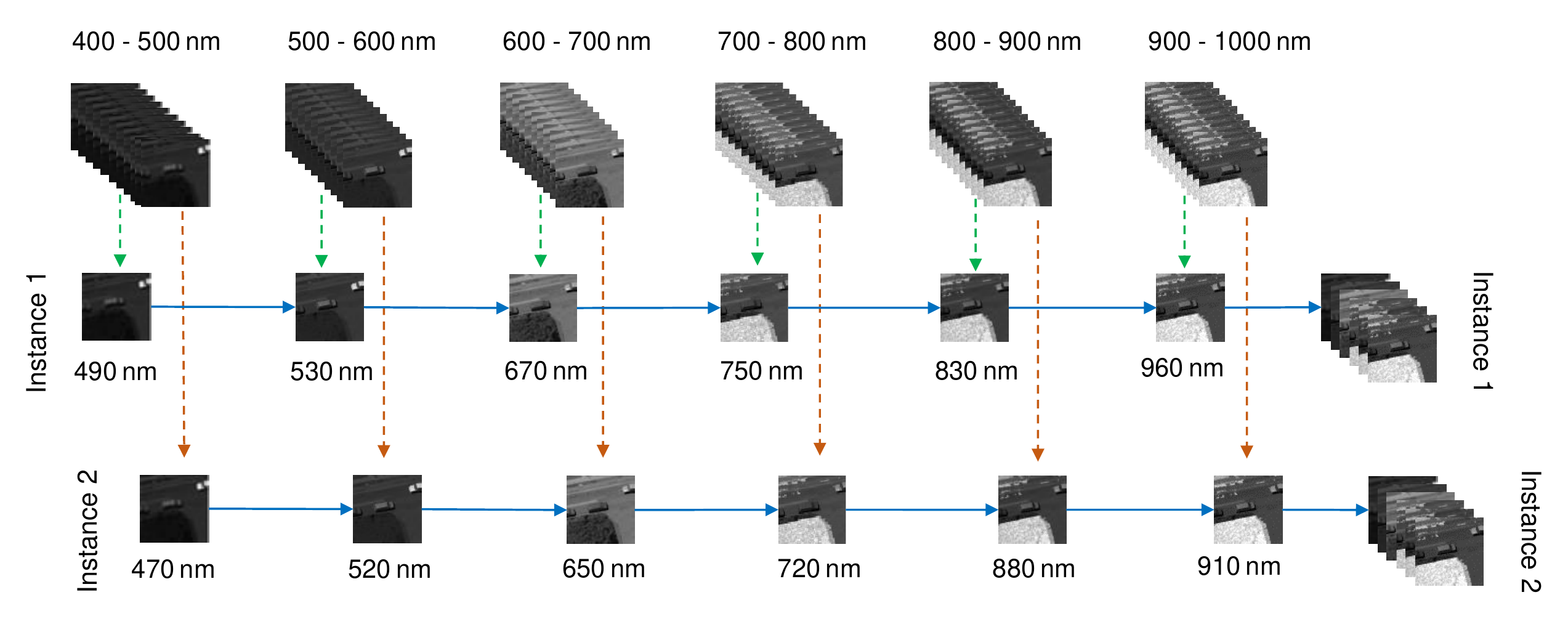}\\
\caption{The proposed hyperspectral data augmentation to increase spectral variance in the dataset. One can observe the variance in the same sample at reasonably distinct wavelengths.}
\label{fig:hsi_augmentation}
\end{figure*}

The data augmentation is highly important in our case as we mimic the WAMI platform in a dataset consisting of fully synthetic images. In particular, it is difficult to approximate the spectral sensitivity curve of a real platform synthetically. The same car samples from different wavelengths are augmented to better approximate the WAMI platform internal mechanics. We stick with 61 channels in the \textit{visible} (400 nm) to \textit{near infrared} (1000 nm) wavelength range. In a single-band image setting with 0.2 fps, we produce about 180 images leading to small spectral variance in the dataset. By using all 61 channels, we generate over 1000 images over the 9 simulations, considering \textit{time} and \textit{spectral} depth. This approach has the potential downside of generating to a dataset dominated by highly similar images. To address this, we sample 6 bands from 6 uniform distributions covering the 61 channels as shown in Fig.~\ref{fig:hsi_augmentation}. This increases the spectral variance while ensuring a reasonably large gap between the augmented images at different wavelengths.

\subsection{Positive and Negative Samples Collection}
The procedure described above produces 27613 vehicle chips ($64\times64$ px) and the vehicles are located in the central position of the positive chips. Similar to the WAMI platform, a vehicle is represented by 20 $\times$ 10 pixels on average in the generated scenarios. Adding context in positive samples seems to improve the learned weights in a CNN \cite{yi2016vehicle}. To collect negative samples, we perform hard-negative mining by considering areas surrounding the positive samples. A negative sample is randomly captured from an area whose center is T =  $10$ - $30$ pixels away from the center of the positive sample. 
\begin{figure}[!h]
\centering
\subfloat[DIRSIG positive samples]{\includegraphics[width=0.23\textwidth]{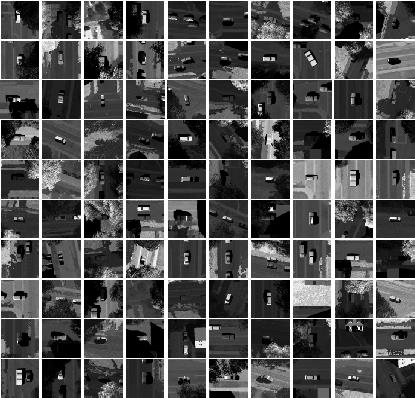}}
\hfill
\subfloat[WAMI positive samples]{\includegraphics[width=0.23\textwidth]{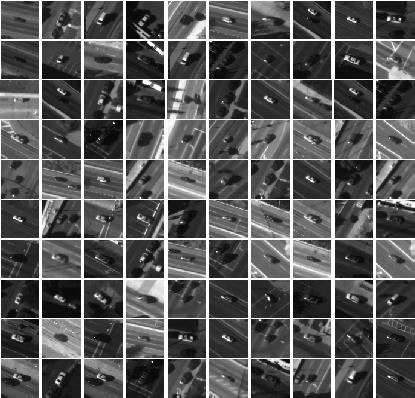}}
\hfill
\caption{Positive training samples collected in the DIRSIG vehicle classification 
dataset and real platform validation samples collected from the WAMI platform 
(\textit{CLIFF06} and \textit{CLIFF07}) videos. The collected chips occupy $64\times64$ pixels in both DIRSIG and WAMI dataset.}
\label{fig:positive_samples}
\end{figure}

Our final dataset consists of 55226 chips captured from different positions of \textit{Mega-Scene I} at different times. To validate the performance on the WAMI platform, we annotate 600 positive and negative chips from the \textit{CLIFF06} and \textit{CLIFF07} videos captured from the WAMI platform. Some of the positive samples from the training and validation dataset can be visualized in Fig.~\ref{fig:positive_samples}. Finally, we train a well-known CNN architecture to perform vehicle classification on the WAMI platform.

\subsection{Training the models}
The architecture used in this study is the ZFNet, an optimized version of AlexNet. We adopt two different training strategies : (1) training from scratch and (2) fine-tuning the weights learned on the ImageNet using the synthetic aerial vehicle detection dataset. In the latter approach, the learning rate is set to 0.0001 other than the classification layer. The classification layer is assigned a learning rate of 0.0005. On the other hand, in the former method, we tune the learning rate to 0.1. The ZFNet from scratch is trained for 4200 iterations with the batch size of 64 whereas the one pre-trained on ImageNet is trained for 400 iterations with the same batch size. In this two experiments, the networks are validated on the 600 samples from WAMI (CLIFF06 and CLIFF07). To integrate further context information into the learned weights, we introduce dilated convolutions with hole size $2$ and $1$ in the \textit{1st} and \textit{2nd} convolutional layers \cite{yu2015multi}. Finally, we follow a two-stage training strategy that uses the $200$ WAMI samples to further update the weights from the fine-tuned ZFNet. The model is validated on the remaining $400$ WAMI samples. This, as expected, boosts the classification accuracy on the WAMI platform. Training is performed on the NVIDIA Tesla K80 GPU in Caffe framework \cite{jia2014caffe}.
 
\begin{table}[h]
\centering
\begin{tabular}{@{}lcccc@{}}
\toprule
Method & \begin{tabular}[c]{@{}c@{}}ZFNet \\ (from Scratch)\end{tabular} & \begin{tabular}[c]{@{}c@{}}ZFNet \\ (ImageNet)\end{tabular} & ZFNet* & \begin{tabular}[c]{@{}c@{}}AlexNet\\ (WAMI \cite{yi2016vehicle})\end{tabular} \\ \midrule \midrule
Accuracy (\%) & 93.20 & 92.230 & 97.0 & 97.1 \\ \bottomrule
\end{tabular}
\caption{Performance of the trained neural networks on the WAMI vehicle classification task. In the "*" case, 200 WAMI samples are used to further fine-tune and validate the network on the 400 WAMI samples. \cite{yi2016vehicle} splits the WPAFB2009 video (WAMI) into training and validation.}
\label{table:Performance_Table_CNN}
\end{table}

As seen in Table~\ref{table:Performance_Table_CNN}, over $93\%$ accuracy is achieved by only using our synthetic dataset to train the ZFNet. This proves the high fidelity of the hyperspectral scenario used to evaluate the deep hyperspectral kernelized correlation filter tracker. To further improve the accuracy, a small amount of WAMI samples are used to train a more WAMI domain-specific model. This improves the accuracy up to $97\%$ reaching the state-of-the-art in vehicle classification in WAMI platform. Finally, with the availability of this dataset, the need to collect a large amount of training samples from the WAMI platform is removed. To support the aerial vehicle detection related studies, we plan on releasing the full images with ground truth locations of the vehicles. This will give more freedom to train detection-domain architectures such as Faster-RCNN \cite{ren2015faster}, R-FCNN \cite{dai2016r}, YOLO9000 \cite{redmon2016yolo9000}, and SSD \cite{liu2016ssd}. 

\section{Tracking Experiments}
\label{section:Tracking Experiments}
\textit{DIRSIG} is a very useful program for generating remote sensing images with high fidelity. This is proved in the previous section where we generated a large synthetic dataset representative of the WAMI platform and trained a convolutional network to classify real images from the WAMI platform. A hyperspectral tracking video representative of the RITMOS sensor was generated in previous studies \cite{uzkent2015efficient, uzkent2017aerial, uzkent2016integrating, uzkent2013feature}. The hyperspectral tracking scenario has two different videos : (1) without trees and (2) with dense trees ($25\%$ full occlusion by trees). Both videos have 1.42 fps and 157 frames with same vehicular traffic, and platform position. For our study, we used both videos to evaluate the performance of the proposed DeepHKCF tracker, its variants, and other hyperspectral state-of-the-art trackers.

\subsection{Hyperparameter Tuning}
This section discusses the hyperparameters that need to be tuned to perform optimal tracking considering both run-time performance and accuracy. The KCF has a number of hyperparameters including padding size, desired Gaussian response width, learning rate and Gaussian kernel bandwith. Our approach has three main hyperparameters: (1) full ROI size, (2) overlap between ROIs and (3) PSR threshold to remove the contribution of noisy response maps. We set the size of a single ROI to (48 $\times$ 48) pixels since each vehicle occupies about 20 $\times$ 10 pixels and hence reasonable content is captured. This removes the need to have a padding size hyperparameter in the DeepHKCF. The other KCF hyperparameters are tuned to similar values as the original KCF paper. The overlap between ROIs is set to $64\%$ in each dimension (Sect.~\ref{overlap_ratio_exps}) whereas the full ROI size and PSR threshold are set to 96 $\times$ 96 pixels (Sect.~\ref{sect:roi_experiments}) and $7$ respectively.

\subsection{Tracking Performance Metrics}

For analyzing the performance of our tracker and its variants, we use two metrics: (1) Central Location Error and (2) Precision, which are defined as follows:

\begin{flushleft}
\textbf{Central Location Error}
\end{flushleft}
The central location error (CLE) for a dataset can be calculated in three effective steps: (1) The central location error is defined as the average Euclidean distance between the predicted center location of the target and the ground truth of a frame. (2) The average center location error over all the frames of one sequence is used to then summarize the overall performance value for that sequence. (3) Lastly, the \emph{average} central location error of a dataset is calculated by averaging the central location error across all the sequences in the dataset. Ideally, it is preferred to have a \emph{low} central location error. 

\begin{flushleft}
\textbf{Precision}
\end{flushleft}
Precision can be defined as thresholding the Euclidean distance between the prediction and ground truth centroid. In the paper, the final Precision scores are obtained by: (1) dividing the number of successful frames to the total number of frames in a sequence to get the Precision score at the respective threshold. (2) Performing the same operation on all the sequences and averaging to compute the final Precision score on a dataset. Pr 20 px and Pr 50 px represent the precision at 20 and 50 pixels Euclidean distance thresholds. The threshold is slided between between $0$ to $50$ pixels by 1 pixel interval to draw the Precision figures. Ideally, it is preferred to have a \emph{high} Precision value. 

\begin{table*}[tbp]
\centering
\resizebox{\textwidth}{!}{%
\begin{tabular}{lccccccccc}
\hline
\multicolumn{1}{|l|}{Method} & \multicolumn{1}{c|}{\begin{tabular}[c]{@{}c@{}}DeepHKCF\\ ZFNet-2\\ (4 x 4 ROI)\end{tabular}} & \multicolumn{1}{c|}{\begin{tabular}[c]{@{}c@{}}DeepHKCF\\ VGG16-5\\ (4 x 4 ROI)\end{tabular}} & \multicolumn{1}{c|}{\begin{tabular}[c]{@{}c@{}}FastDeepHKCF\\ VGG16-5\\ (4 x 4 ROI)\end{tabular}} & \multicolumn{1}{c|}{\begin{tabular}[c]{@{}c@{}}HKCF\\ HSI + fHOG\\ (4 x 4 ROI)\end{tabular}} & \multicolumn{1}{c|}{\begin{tabular}[c]{@{}c@{}}HKCF\\ HSI\\ (4 x 4 ROI)\end{tabular}} & \multicolumn{1}{c|}{\begin{tabular}[c]{@{}c@{}}HKCF\\ fHOG\\ (4 x 4 ROI)\end{tabular}} & \multicolumn{1}{c|}{\begin{tabular}[c]{@{}c@{}}HKCF\\ fHOG\\ (1 x 1 ROI)\end{tabular}} & \multicolumn{1}{c|}{\begin{tabular}[c]{@{}c@{}}ECO\\ fHOG\end{tabular}} & \multicolumn{1}{c|}{\begin{tabular}[c]{@{}c@{}}ECO\\ VGG16-5\end{tabular}} \\ \hline
\hline
\multicolumn{1}{|l|}{\begin{tabular}[c]{@{}l@{}}Pr. (20 px)\end{tabular}} & \multicolumn{1}{c|}{\color{green}70.13} & \multicolumn{1}{c|}{\color{red}68.45} & \multicolumn{1}{c|}{\color{blue}66.26} & \multicolumn{1}{c|}{38.79} & \multicolumn{1}{c|}{39.53} & \multicolumn{1}{c|}{39.30} & \multicolumn{1}{c|}{38.74} & \multicolumn{1}{c|}{39.86} & \multicolumn{1}{c|}{64.15} \\ \hline
\multicolumn{1}{|l|}{\begin{tabular}[c]{@{}l@{}}Pr. (50 px)\end{tabular}} & \multicolumn{1}{c|}{\color{green}81.05} & \multicolumn{1}{c|}{\color{blue}80.27} & \multicolumn{1}{c|}{\color{red}80.65} & \multicolumn{1}{c|}{57.56} & \multicolumn{1}{c|}{54.30} & \multicolumn{1}{c|}{58.58} & \multicolumn{1}{c|}{42.12} & \multicolumn{1}{c|}{43.24} & \multicolumn{1}{c|}{67.61} \\ \hline
\multicolumn{1}{|l|}{\begin{tabular}[c]{@{}l@{}}CLE\end{tabular}} & \multicolumn{1}{c|}{\color{green}48.97} & \multicolumn{1}{c|}{\color{blue}51.71} & \multicolumn{1}{c|}{\color{red}51.15} & \multicolumn{1}{c|}{118.73} & \multicolumn{1}{c|}{146.30} & \multicolumn{1}{c|}{119.04} & \multicolumn{1}{c|}{179.71} & \multicolumn{1}{c|}{168.43} & \multicolumn{1}{c|}{113.46} \\ \hline
\multicolumn{1}{|l|}{\begin{tabular}[c]{@{}l@{}}FPS\end{tabular}} & \multicolumn{1}{c|}{0.51} & \multicolumn{1}{c|}{0.22} & \multicolumn{1}{c|}{1.11} & \multicolumn{1}{c|}{\color{red}3.01} & \multicolumn{1}{c|}{2.32} & \multicolumn{1}{c|}{\color{blue}2.98} & \multicolumn{1}{c|}{\color{green}25.11} & \multicolumn{1}{c|}{2.70} & \multicolumn{1}{c|}{1.19} \\ \hline
\end{tabular}%
}
\color{green}--- Best \hspace{.15\linewidth}\color{red}--- $2^{nd}$ Best \hspace{.15\linewidth}\color{blue}--- $3^{th}$ Best \color{black}
\caption{A detailed analysis of the DeepHKCF tracker, its variants and original KCF and ECO trackers on the \textit{DIRSIG} \textit{no-trees} video. The DeepHKCF tracker with ROI mapping delivers the optimal results considering the trade-off between run-time and tracking performance. The experiments are carried out on a CPU with 8GB RAM and 2.9GHz i5 processor and the \textit{CLE} and \textit{FPS} represent the average central location error of the tracker and operation frame rate per second.}
\label{tab:detailed_results_no_trees}
\end{table*}
\begin{table*}[tbp]
\centering
\resizebox{\textwidth}{!}{%
\begin{tabular}{lccccccccc}
\hline
\multicolumn{1}{|l|}{Method} & \multicolumn{1}{c|}{\begin{tabular}[c]{@{}c@{}}DeepHKCF\\ ZFNet-2\\ (4 x 4 ROI)\end{tabular}} & \multicolumn{1}{c|}{\begin{tabular}[c]{@{}c@{}}DeepHKCF\\ VGG16-5\\ (4 x 4 ROI)\end{tabular}} & \multicolumn{1}{c|}{\begin{tabular}[c]{@{}c@{}}FastDeepHKCF\\ VGG16-5\\ (4 x 4 ROI)\end{tabular}} & \multicolumn{1}{c|}{\begin{tabular}[c]{@{}c@{}}HKCF\\ HSI + fHOG\\ (4 x 4 ROI)\end{tabular}} & \multicolumn{1}{c|}{\begin{tabular}[c]{@{}c@{}}HKCF\\ HSI\\ (4 x 4 ROI)\end{tabular}} & \multicolumn{1}{c|}{\begin{tabular}[c]{@{}c@{}}HKCF\\ fHOG\\ (4 x 4 ROI)\end{tabular}} & \multicolumn{1}{c|}{\begin{tabular}[c]{@{}c@{}}HKCF\\ fHOG\\ (1 x 1 ROI)\end{tabular}} & \multicolumn{1}{c|}{\begin{tabular}[c]{@{}c@{}}ECO\\ fHOG\end{tabular}} & \multicolumn{1}{c|}{\begin{tabular}[c]{@{}c@{}}ECO\\ VGG16-5\end{tabular}} \\ \hline
\hline
\multicolumn{1}{|l|}{\begin{tabular}[c]{@{}l@{}}Pr. (20 px)\end{tabular}} & \multicolumn{1}{c|}{\color{red}38.08} & \multicolumn{1}{c|}{\color{blue}37.48} & \multicolumn{1}{c|}{31.68} & \multicolumn{1}{c|}{23.57} & \multicolumn{1}{c|}{23.88} & \multicolumn{1}{c|}{24.69} & \multicolumn{1}{c|}{17.31} & \multicolumn{1}{c|}{25.51} & \multicolumn{1}{c|}{\color{green}40.07} \\ \hline
\multicolumn{1}{|l|}{\begin{tabular}[c]{@{}l@{}}Pr. (50 px)\end{tabular}} & \multicolumn{1}{c|}{\color{blue}43.83} & \multicolumn{1}{c|}{\color{green}44.16} & \multicolumn{1}{c|}{\color{red}44.01} & \multicolumn{1}{c|}{33.80} & \multicolumn{1}{c|}{32.82} & \multicolumn{1}{c|}{36.28} & \multicolumn{1}{c|}{25.33} & \multicolumn{1}{c|}{28.81} & \multicolumn{1}{c|}{42.15} \\ \hline
\multicolumn{1}{|l|}{\begin{tabular}[c]{@{}l@{}}CLE\end{tabular}} & \multicolumn{1}{c|}{\color{blue}156.77} & \multicolumn{1}{c|}{\color{red}156.67} & \multicolumn{1}{c|}{\color{green}143.66} & \multicolumn{1}{c|}{196.80} & \multicolumn{1}{c|}{191.19} & \multicolumn{1}{c|}{180.47} & \multicolumn{1}{c|}{209.63} & \multicolumn{1}{c|}{222.61} & \multicolumn{1}{c|}{181.64} \\ \hline
\multicolumn{1}{|l|}{\begin{tabular}[c]{@{}l@{}}FPS\end{tabular}} & \multicolumn{1}{c|}{0.41} & \multicolumn{1}{c|}{0.17} & \multicolumn{1}{c|}{0.65} & \multicolumn{1}{c|}{1.67} & \multicolumn{1}{c|}{2.10} & \multicolumn{1}{c|}{\color{red}2.59} & \multicolumn{1}{c|}{\color{green}8.22} & \multicolumn{1}{c|}{\color{blue}2.55} & \multicolumn{1}{c|}{0.83} \\ \hline
\end{tabular}%
}
\color{green}--- Best \hspace{.15\linewidth}\color{red}--- $2^{nd}$ Best \hspace{.15\linewidth}\color{blue}--- $3^{th}$ Best \color{black}
\caption{A detailed analysis of the DeepHKCF tracker, its variants and original KCF and ECO trackers on the \textit{DIRSIG} \textit{dense trees} video.}
\label{tab:detailed_results_trees}
\end{table*}
\subsection{Results on the No-trees Scenario}
After tuning the hyperparameters of DeepHKCF, we test it on the 43 vehicles in the \textit{no-trees} scenario. We compare the DeepHKCF to a number of variants of the proposed single KCF multiple ROIs approach as seen in Table~\ref{tab:detailed_results_no_trees}. Furthermore, we perform experiments on the original KCF algorithm (single KCF-single ROI approach) with the same ROI size ($48$ $\times$ $48$ px). Additionally, we compare the proposed tracker to Efficient Convolutional Operator tracker (ECO) \cite{danelljan2016eco} that is ranked first in the VOT16 tracking benchmark. For fair comparison, we increase the learning rate of the ECO tracker to match our scenarios and use the same features. Similar to DeepHKCF and HKCF, we use the activations of the $5th$ convolutional layer of the VGGNet and fHoG features. To determine the ROI area, ECO considers the padding size of 3.5, larger than the optimal padding size (2.0) of the KCF. We keep this hyper-parameter same as including further background deteriorates the features. For $20\times10$ pixel vehicle, the ECO and vanilla-form KCF trackers have a search area of $80\times45$, and $60\times30$ pixels whereas it is $96\times96$ for the proposed DeepHKCF. Fig.~\ref{fig:precision_notrees} and Table~\ref{tab:detailed_results_no_trees} show the performances of the proposed DeepHKCF, its variants and the baseline KCF and ECO trackers.

The DeepHKCF performs exceptionally well in the no-trees scenario, achieving $70\%$ precision at the 20 px threshold and outperforming all the baseline methods by a large margin. Meanwhile, the proposed HKCF with fHoG features performs substantially worse than the one with deep features. However, it outruns the original KCF with fHoG features by a large margin at $50$ px precision as shown in Table~\ref{tab:detailed_results_no_trees}, proving the contribution of the proposed single KCF multiple ROIs approach in low frame rate tracking. Concatenating hyperspectral features with fHoG slightly degrades the precision whereas hyperspectral feature channels alone performs worse than the former methods. This indicates that the NIR channels do not contribute to tracking in the KCF framework. The ECO tracker, on the other hand, delivers $10$ - $15\%$ lower accuracy than the DeepHKCF trackers at 20 px precision and about $20\%$ worse in terms of the precision at 50 px and central location error. All in all, the DeepHKCF tracker with ROI mapping (FastDeepHKCF) achieves optimal results considering its reasonably high tracking accuracy and highest operation rate among the DeepHKCF trackers.

\subsection{Results on the Dense Trees Scenario}
\begin{figure}[t]
\centering
\subfloat[Precision curve - with trees scenario]{
	\centering
    \includegraphics[width=0.75\linewidth]{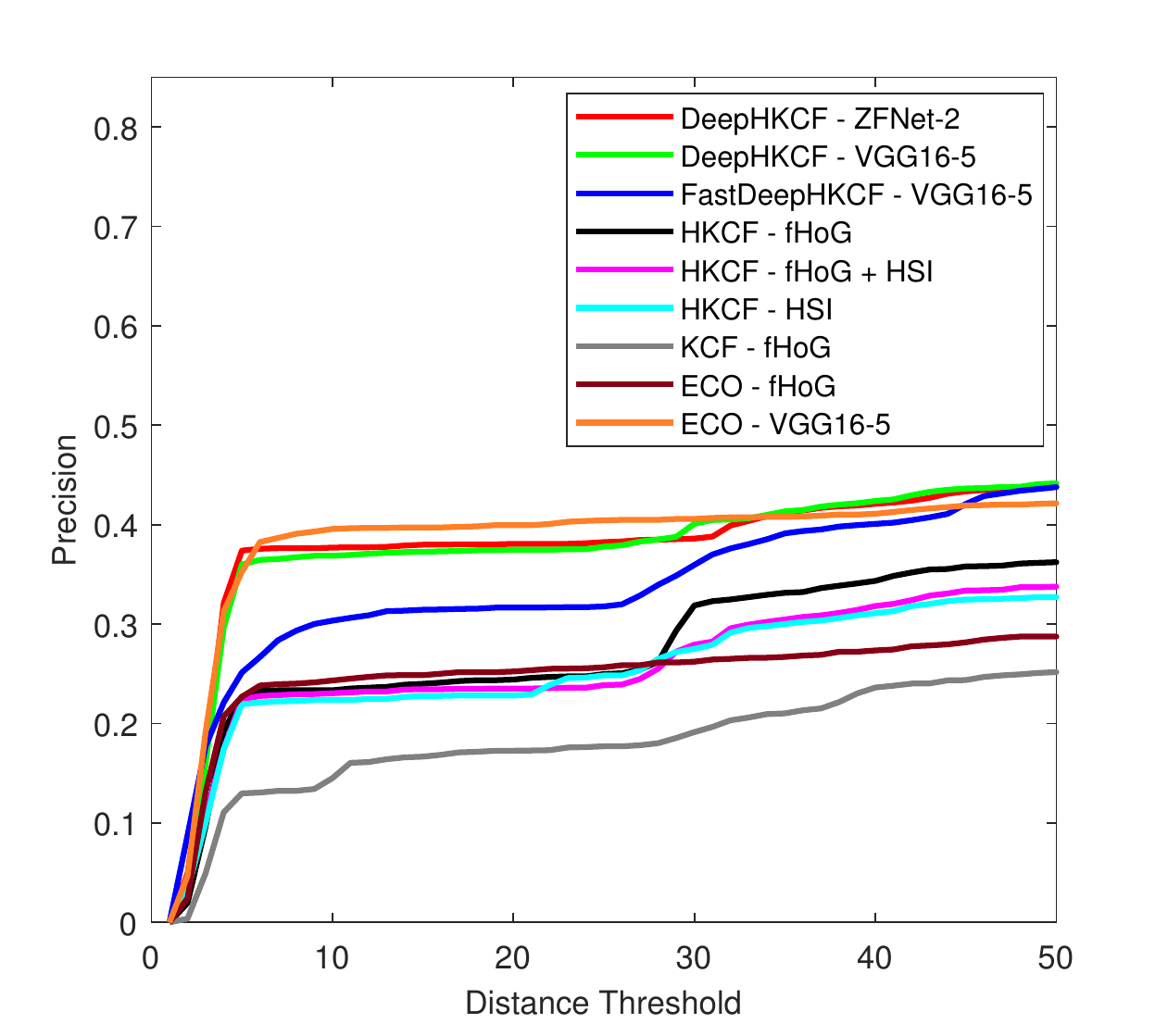}
    \label{fig:precision_trees}
    }
\\
\subfloat[Precision curve - without trees scenario]
	{
    \centering
    \includegraphics[width=0.75\linewidth]{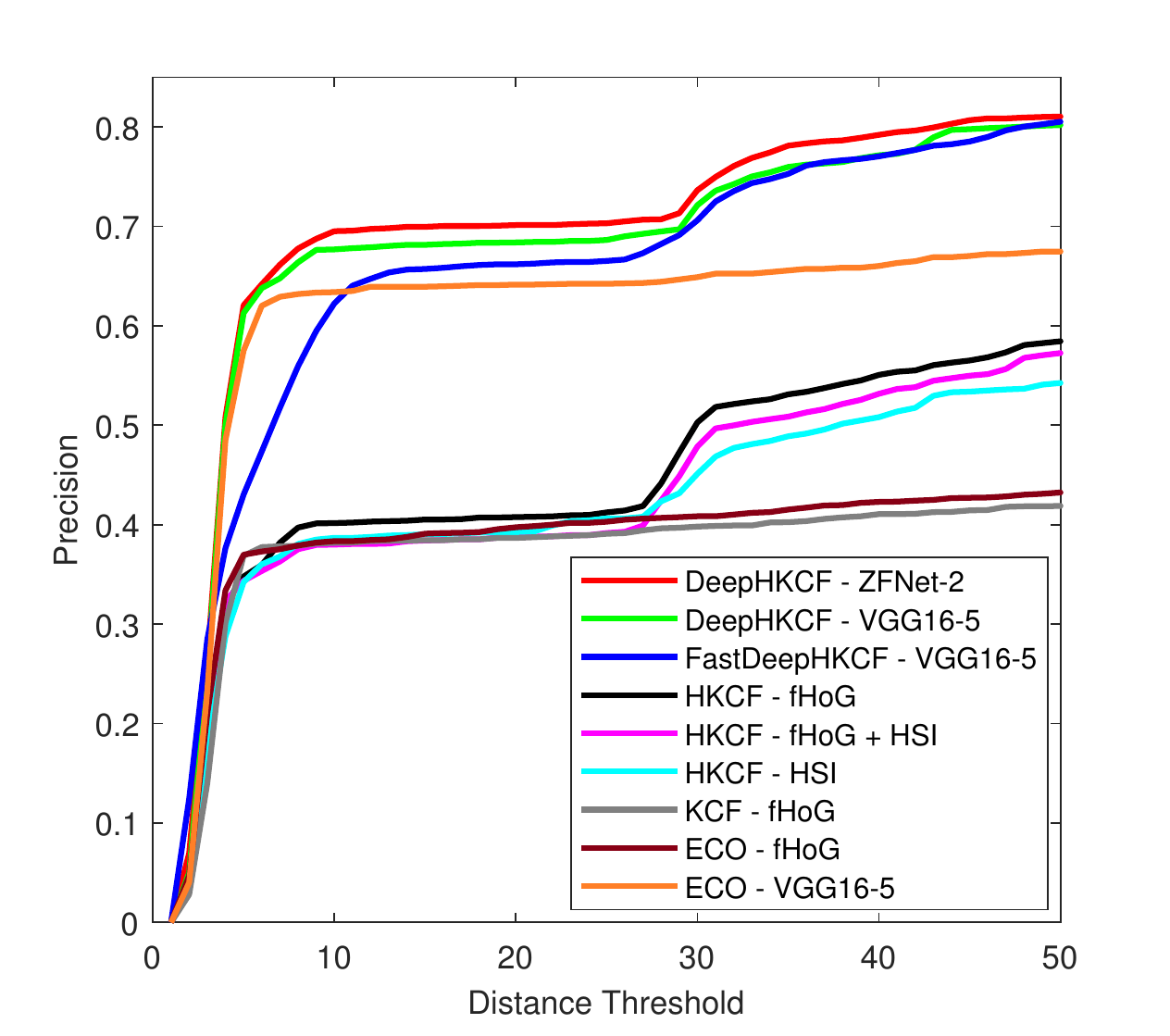}
    \label{fig:precision_notrees}}
\hfill
\caption{Precision curves of the DeepHCKF tracker and its variants on the \textit{DIRSIG} scenario with trees and without trees. The ZFNet-2 and VGGNet-5 refer to the activations of the \textit{2nd} and \textit{5th} convolutional layers.}
\label{fig:precision}
\end{figure}

In addition to conducting experiments on the no-trees scenario, we run the DeepHKCF tracker and its variants on the same scenario with \textit{dense trees}. This is an extremely challenging scene dominated by large trees and their shadows as shown in Fig.~\ref{fig:dirsig_scenarios}. On average, a vehicle is fully occluded in 1 out of 4 frames. Severe occlusions combined with low frame rate make this a more challenging scene. The DeepHKCF trackers outperform the other baseline methods other than ECO tracker by a large margin as in the no-trees scenario (Fig.~\ref{fig:precision_notrees}). At 20 px precision, the DeepHKCF tracker achieves about $39\%$ accuracy whereas others perform $10$-$20\%$ worse. On the other hand, among the DeepHKCF trackers, the FastDeepHKCF, delivers similar precision at 50 px and higher frame rate. Similar to the no trees scenario, the combination of hyperspectral feature channels with fHoG degrades the performance with respect to the fHoG-only features. We believe that this could be due to more frequent switching to non-vehicle objects with similar hyperspectral features to the target of interest through occlusions. By using fHoG-only features, it is less likely to switch to an object that does not appear like a vehicle.
The dramatic drop in precision rates between the no-trees and dense trees scenarios is easily seen in Fig.~\ref{fig:precision}. This is likely due to three major reasons : (1) high frequency of severe occlusions, (2) low video frame rate, and (3) relatively smaller search area considered by our single KCF-multiple ROIs approach. The combination of the first two reasons leads to dramatically large displacement of objects in between the frames where they are visible. This results in the targets being located out of the search area of the DeepHKCF tracker. There are two solutions to address the challenge of tracking through severe occlusions. The first and less practical solution is increasing the full ROI size in each dimension. This way, we increase the likelihood of keeping the target in our search area traveling through severe occlusions. However, this will also reduce the run-time performance. A more practical solution could be delivered by leveraging a Bayes Filter. For instance, \cite{uzkent2017aerial,uzkent2015efficient} uses a Bayes Filter in a Multi-dimensional Assignment algorithm to update the measurements in light of the later measurements in the same scenario. This way, we can low-pass the unlikely jumps that occurs during severe occlusions. On the other hand, we believe that, increasing the search area in a practical manner might be the key to achieving state-of-the-art performance in scenarios dominated by trees (see Sect.~\ref{sect:roi_experiments}).

\begin{table}[b]
\centering
\resizebox{\linewidth}{!}{%
\begin{tabular}{lccc}
\hline
\multicolumn{1}{|l|}{Method} & \multicolumn{1}{c|}{\begin{tabular}[c]{@{}c@{}}DeepHCKF\\ VGGNet-5\end{tabular}} & \multicolumn{1}{c|}{\begin{tabular}[c]{@{}c@{}}FastDeepHKCF\\ VGGNet-5\end{tabular}} & \multicolumn{1}{c|}{\begin{tabular}[c]{@{}c@{}}ECO\\ VGGNet-5\end{tabular}} \\ \hline \hline
\multicolumn{1}{|l|}{\begin{tabular}[c]{@{}l@{}}Pr. (20 px)\end{tabular}} & \multicolumn{1}{c|}{\textbf{44.34}} & \multicolumn{1}{c|}{39.11} & \multicolumn{1}{c|}{29.79} \\ \hline
\multicolumn{1}{|l|}{\begin{tabular}[c]{@{}l@{}}Pr. (50 px)\end{tabular}} & \multicolumn{1}{c|}{\textbf{53.30}} & \multicolumn{1}{c|}{47.67} & \multicolumn{1}{c|}{34.29} \\ \hline
\multicolumn{1}{|l|}{CLE} & \multicolumn{1}{c|}{\textbf{156.98}} & \multicolumn{1}{c|}{174.53} & \multicolumn{1}{c|}{204.85} \\ \hline
\multicolumn{1}{|l|}{FPS} & \multicolumn{1}{c|}{0.51} & \multicolumn{1}{c|}{1.11} & \multicolumn{1}{c|}{\textbf{1.12}} \\ \hline
\end{tabular}%
}
\caption{Performance of the DeepHKCF and ECO trackers on the temporally down-sampled no-trees video (0.7 fps).}
\label{table:comparison_downsampled_video}
\end{table}

\subsection{Experiments on Temporally Down-sampled Video}
In the previous sections, we evaluated the proposed trackers and the baseline methods on the 1.42 fps videos with dense trees and without trees. The ECO tracker performs only slightly worse to the DeepHKCF trackers at 20 px precision. We believe that this might be due to slowly moving or stopped vehicles. To further test their performance with respect to more drastic target motion, we down-sample the \textit{video without trees} temporally by \textit{two}, resulting in 0.7 fps video. All the hyperparameters in the FastDeepHKCF and ECO are kept same as before for fair comparison. As seen in Table~\ref{table:comparison_downsampled_video}, the DeepHKCF trackers outperform the ECO tracker by a large margin in terms of precision and central location error showing its robustness to extreme target displacement in successive frames. The ECO tracker misses more targets due to the smaller ROI considered in the detection operation. 

\subsection{Comparison with Hyperspectral Trackers}

In the previous section the proposed tracker is compared to the state-of-the-art discriminative trackers. In this section, we compare the proposed DeepHKCF tracker to the generative hyperspectral trackers \cite{uzkent2016real,uzkent2017aerial}. These trackers are designed for the \textit{DIRSIG} scenarios and extensively use the multi-dimensional assignment algorithm (MDA) \cite{poore1994multidimensional}. HFT \cite{uzkent2016real} relies on off-line trained road and car classifiers to optimize the search space. The hyperspectral histograms are then computed in a sliding window to assign similarity scores. The obtained heat map is then thresholded and post-processed to find the blobs, which are assigned to the target using MDA. The HLT \cite{uzkent2017aerial}, on the other hand, learns a generative target model using hyperspectral likelihood maps rather than using off-line trained classifiers. The blobs are extracted from the final thresholded map and track statistics are updated from the past N frames using the MDA. Here, we compare DeepHKCF only with HLT in Table~\ref{table:Comparison_to_other_trackers} since HFT relies on car and asphalt classifiers trained on the samples from the same scene.

\begin{table}[h]
\centering
\resizebox{0.90\linewidth}{!}{%
\begin{tabular}{|l|c|c|c|c|}
\hline
Method & \begin{tabular}[c]{@{}c@{}}DeepHKCF\\ ZFNet-2\end{tabular} & \begin{tabular}[c]{@{}c@{}}FastDeepHKCF\\ VGGNet-5\end{tabular} & HLT \cite{uzkent2017aerial} (5D) & HLT \cite{uzkent2017aerial} (2D) \\  \hline \hline
\begin{tabular}[c]{@{}l@{}}Pr. (20 px)\\ Trees\end{tabular} & 38.08 & 31.71 & \textbf{51.69} & 41.86 \\ \hline 
\begin{tabular}[c]{@{}l@{}}Pr. (20 px) \\ No trees\end{tabular} & \textbf{70.13} & 66.26 & 64.42 & 57.25 \\ \hline
\begin{tabular}[c]{@{}l@{}}Pr. (50 px)\\ Trees\end{tabular} & 43.83 & 44.01 & \textbf{55.12} & 46.72 \\ \hline
\begin{tabular}[c]{@{}l@{}}Pr. (50 px) \\ No trees\end{tabular} & \textbf{81.05} & 80.27 & 71.27 & 68.31 \\ \hline
\begin{tabular}[c]{@{}l@{}}CLE \\ Trees\end{tabular} & 156.74 & 143.66 & \textbf{135.03} & 158.12 \\ \hline
\begin{tabular}[c]{@{}l@{}}CLE\\ No trees\end{tabular} & \textbf{48.97} & 51.71 & 65.36 & 91.97 \\ \hline
FPS & 0.51 & \textbf{1.11} & 1.01 & 1.09 \\ \hline
\end{tabular}%
}
\caption{Comparison of the DeepHKCF tracker to HLT. The $5D$ and $2D$ refer to the number of past frames considered in MDA \cite{poore1994multidimensional}. The experiments are carried out on a CPU with 8GB RAM and 2.9GHz i5 processor. The best result in each category has been highlighted for better understanding.}
\label{table:Comparison_to_other_trackers}
\end{table}

As seen in Table~\ref{table:Comparison_to_other_trackers}, the use of a Bayes Filter and the multi-dimensional assignment algorithm (MDA) is crucial in a scenario largely dominated by occlusions. We can see the effect of reducing the length of the time window in MDA as the HLT's performance drops drastically by reducing the width from $5$-D to $2$-D, especially for the scenario with \textit{trees}. The proposed DeepHKCF trackers outperform HLT in the no-trees scenario by about $30\%$ in terms of central location error, thus establishing its dominance in a scenario without occlusion. Finally, the FastDeepHKCF delivers the optimal results considering the trade-off between tracking accuracy and run-time performance.

\subsection{Effect of Overlap Ratio}
\label{overlap_ratio_exps}
\begin{figure}[!h]
\centering
\subfloat[Varying overlap ratio between ROIs in FastDeepHKCF]{\includegraphics[width=0.85\linewidth]{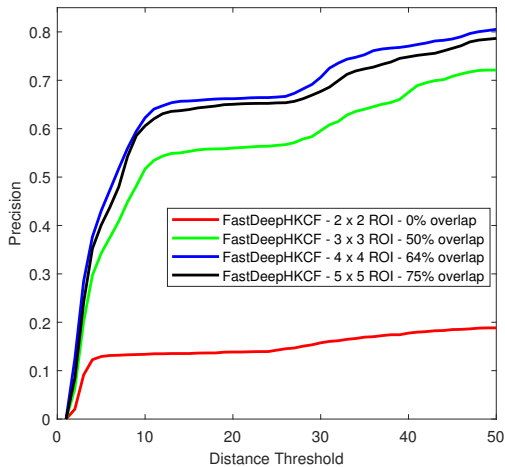}\label{fig:number_rois_effect}}
\\
\subfloat[Varying full ROI size in DeepHKCF]{\includegraphics[width=0.85\linewidth]{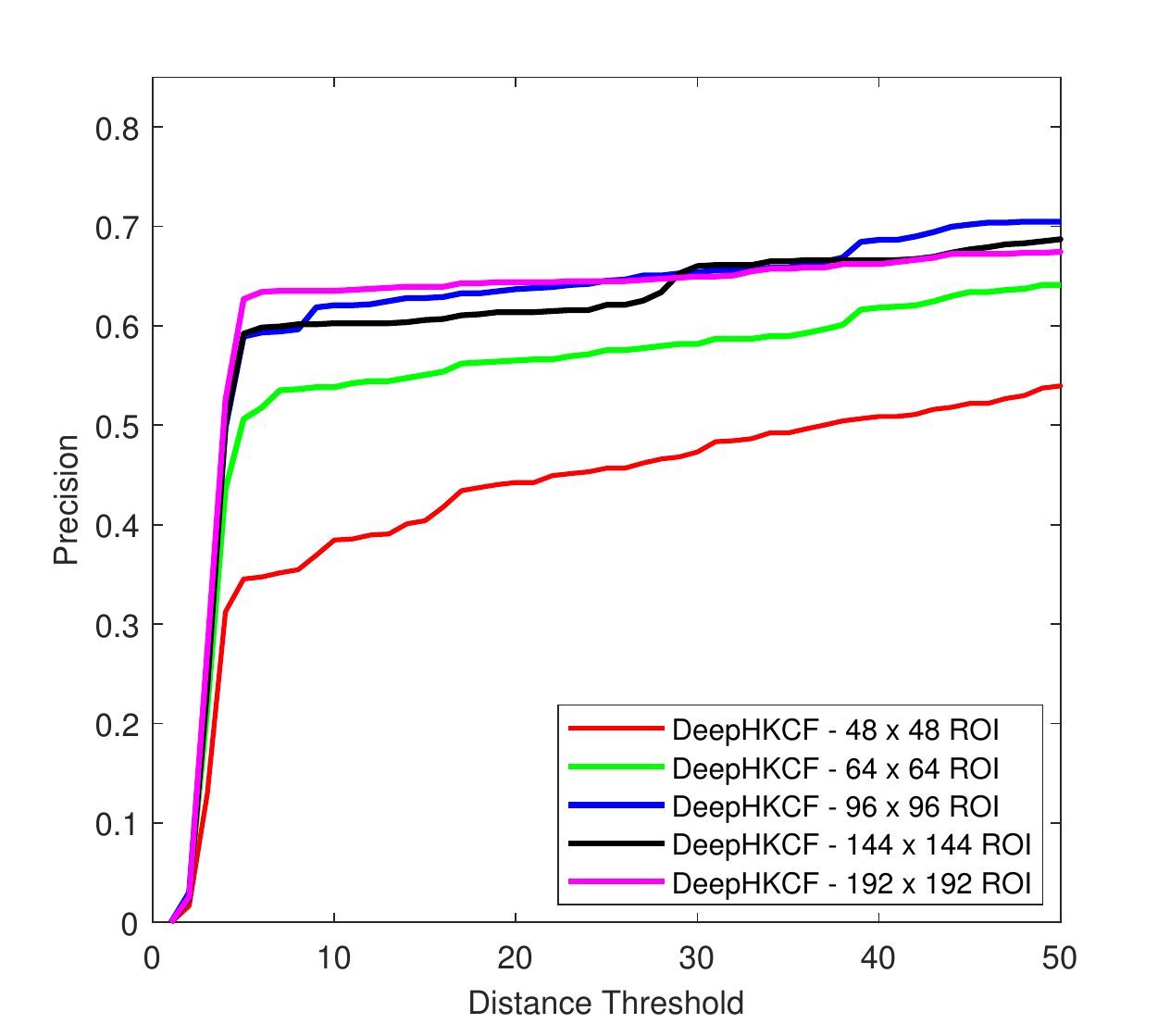}\label{fig:roi_size_effect}}
\hfill
\caption{The FastDeepHKCF tracker is experimented on the \textit{no-trees} scenario with varying overlap ratio in the single KCF-multiple ROIs approach (a) whereas (b) shows the results on the effect of different full ROI sizes in DeepHKCF. In (b), the trackers are run through the first \textit{25 frames} of each target on the \textit{scenario with trees}.}
\label{fig:hyperparameters}
\end{figure}

We experiment on the DeepHKCF tracker with ROI mapping (FastDeepHKCF) as a function of the \textit{overlap ratio} between the adjacent ROIs. As mentioned before, it is necessary to have overlap between the adjacent ROIs as the hanning window is applied to the features before the FFT operation. The hanning window filters the noise at the boundaries resulting from FFT operation. Increasing overlap ratio, at the same time, leads to increased complexity ($O(m*nlog(n))$) due to a larger number of ROIs ($m$) in the full ROI ($96$ $\times$ $96$ px). Fig.~\ref{fig:number_rois_effect} shows the precision rates of FastDeepHKCF tracker with different overlap ratios between the adjacent ROIs.

The $64\%$ and $75\%$ overlap ratios lead to drastically better results than the lower ones (Fig.~\ref{fig:number_rois_effect}). Considering the accuracy/speed trade-off, $64\%$ overlap (m = $16$) is used as the optimal setting.
\subsection{Effect of ROI Size}
\label{sect:roi_experiments}
The overlap between the adjacent ROIs is an essential part of the DeepHKCF tracker as it ensures consideration of the every single point in the full ROI by the correlation filter. Another key parameter in this direction is the full ROI size since we have a low temporal resolution and occlusion-dominated scene. We enlarge the ROI size of the optimal DeepHCKF tracker and observe the performance in the \textit{scene with trees}. The results are shown in Fig.~\ref{fig:roi_size_effect}. In this experiment, the run-time performance of the tracker is ignored as the goal is to measure the contribution of full ROI size.

As shown in Fig.~\ref{fig:roi_size_effect}, the larger ROI size with the same overlap ratio does not necessarily lead to better performance while quadratically increasing the speed. This could be due to growing confusion as the larger ROIs contain a higher number of similar objects. Additionally, these results  demonstrate the obvious need to couple the tracking-by-detection algorithms to a Multi-dimensional Assignment algorithm in a Bayes Filter framework in occlusion-dominated scenes \cite{uzkent2017aerial,uzkent2016real,uzkent2015efficient}.

\section{Conclusion}
Adaptive multi-modal sensors are becoming increasingly important in the aerial tracking domain due to the unique challenges posed by this platforms. In this study, we propose a tracking-by-detection algorithm driven tracker inspired by a multi-modal sensor and deep features. This approach replaces the traditional template-matching based hyperspectral trackers with a new state-of-the-art tracker becoming increasingly popular in traditional visual object tracking. More specifically, we delivered a new framework to handle low temporal resolution in aerial platforms in KCF tracker, called \textit{single KCF-multiple ROIs approach}. To further boost the tracking accuracy, we replaced the traditional features with deep CNN features. Finally, an ROI mapping approach was proposed to speed up extracting features in a single KCF-multiple ROIs approach. The proposed DeepHKCF tracker was evaluated on synthetic scenarios generated by \textit{DIRSIG} software. In the scenario with no-trees, the DeepHKCF tracker performs exceptionally well with $80\%$ precision at 50 px, outperforming other trackers. In the same scenario but dominated by occlusions, it is outperformed by trackers employing a \textit{multi-dimensional assignment algorithm} and \textit{Bayes Filter}. To prove the high-fidelity of the DIRSIG generated scenarios, we build a synthetic, aerial vehicle classification dataset to perform classification on the real-platform (\textit{WAMI}). Our dataset, consisting of $55226$ samples, was used to train CNNs to perform binary classification. We achieve about $93.2\%$ on the WAMI samples by only training on synthetic dataset. This dataset can be highly beneficial in aerial detection and tracking due to limited amount of publicly available data in those domains.

In future work, we plan on supporting the DeepHKCF tracker by integrating a multi-dimensional assignment algorithm and Bayes Filter to better handle severe occlusions.

\ifCLASSOPTIONcaptionsoff
  \newpage
\fi


%

\section*{ACKNOWLEDGEMENTS}
This work has been supported by the Dynamic Data Driven Applications Systems Program, Air Force Office of
Scientific Research under Grant FA9550-11-1-0348.
{\small

\bibliographystyle{IEEEtran}
\bibliography{egbib}

\begin{thebibliography}{10}
\providecommand{\url}[1]{#1}
\csname url@samestyle\endcsname
\providecommand{\newblock}{\relax}
\providecommand{\bibinfo}[2]{#2}
\providecommand{\BIBentrySTDinterwordspacing}{\spaceskip=0pt\relax}
\providecommand{\BIBentryALTinterwordstretchfactor}{4}
\providecommand{\BIBentryALTinterwordspacing}{\spaceskip=\fontdimen2\font plus
\BIBentryALTinterwordstretchfactor\fontdimen3\font minus
  \fontdimen4\font\relax}
\providecommand{\BIBforeignlanguage}[2]{{%
\expandafter\ifx\csname l@#1\endcsname\relax
\typeout{** WARNING: IEEEtran.bst: No hyphenation pattern has been}%
\typeout{** loaded for the language `#1'. Using the pattern for}%
\typeout{** the default language instead.}%
\else
\language=\csname l@#1\endcsname
\fi
#2}}
\providecommand{\BIBdecl}{\relax}
\BIBdecl

\bibitem{pelapur2012persistent}
R.~Pelapur, S.~Candemir, F.~Bunyak, M.~Poostchi, G.~Seetharaman, and
  K.~Palaniappan, ``Persistent target tracking using likelihood fusion in
  wide-area and full motion video sequences,'' in \emph{Information Fusion
  (FUSION), 2012 15th International Conference on}.\hskip 1em plus 0.5em minus
  0.4em\relax IEEE, 2012, pp. 2420--2427.

\bibitem{portmann2014people}
J.~Portmann, S.~Lynen, M.~Chli, and R.~Siegwart, ``People detection and
  tracking from aerial thermal views,'' in \emph{Robotics and Automation
  (ICRA), 2014 IEEE International Conference on}.\hskip 1em plus 0.5em minus
  0.4em\relax IEEE, 2014, pp. 1794--1800.

\bibitem{danelljan2016eco}
M.~Danelljan, G.~Bhat, F.~S. Khan, and M.~Felsberg, ``Eco: Efficient
  convolution operators for tracking,'' \emph{arXiv preprint arXiv:1611.09224},
  2016.

\bibitem{uzkent2015efficient}
B.~Uzkent, M.~J. Hoffman, and A.~Vodacek, ``Efficient integration of spectral
  features for vehicle tracking utilizing an adaptive sensor,'' in
  \emph{IS\&T/SPIE Electronic Imaging}, 2015, pp. 940\,707--940\,707.

\bibitem{uzkent2016real}
B.~Uzkent, \emph{Real-time Aerial Vehicle Detection and Tracking using a
  Multi-modal Optical Sensor}.\hskip 1em plus 0.5em minus 0.4em\relax Rochester
  Institute of Technology, 2016.

\bibitem{uzkent2017aerial}
B.~Uzkent, A.~Rangnekar, and M.~J. Hoffman, ``Aerial vehicle tracking by
  adaptive fusion of hyperspectral likelihood maps,'' in \emph{Computer Vision
  and Pattern Recognition Workshops (CVPRW), 2017 IEEE Conference on}.\hskip
  1em plus 0.5em minus 0.4em\relax IEEE, 2017, pp. 233--242.

\bibitem{WPAFB}
AFRL, ``Wright-patterson air force basevvi (wpafb) dataset,''
  \url{https://www.sdms.afrl.af.mil/index.php?collection=wpafb2009}, 2009.

\bibitem{CLIF}
------, ``Wami columbus large image format (clif) dataset,''
  \url{https://www.sdms.afrl.af.mil/index.php? collection=clif2007}, 2007.

\bibitem{russakovsky2015imagenet}
O.~Russakovsky, J.~Deng, H.~Su, J.~Krause, S.~Satheesh, S.~Ma, Z.~Huang,
  A.~Karpathy, A.~Khosla, M.~Bernstein \emph{et~al.}, ``Imagenet large scale
  visual recognition challenge,'' \emph{International Journal of Computer
  Vision}, vol. 115, no.~3, pp. 211--252, 2015.

\bibitem{uzkent2013feature}
B.~Uzkent, M.~J. Hoffman, A.~Vodacek, J.~P. Kerekes, and B.~Chen, ``Feature
  matching and adaptive prediction models in an object tracking dddas,''
  \emph{Procedia Computer Science}, vol.~18, pp. 1939--1948, 2013.

\bibitem{uzkent2015feature}
B.~Uzkent, M.~J. Hoffman, A.~Vodacek, and B.~Chen, ``Feature matching with an
  adaptive optical sensor in a ground target tracking system,'' \emph{IEEE
  Sensors Journal}, vol.~15, no.~1, pp. 510--519, 2015.

\bibitem{meyer2004ritmos}
R.~D. Meyer, K.~J. Kearney, Z.~Ninkov, C.~T. Cotton, P.~Hammond, and B.~D.
  Statt, ``{RITMOS}: a micromirror-based multi-object spectrometer,'' in
  \emph{SPIE Astronomical Telescopes+ Instrumentation}.\hskip 1em plus 0.5em
  minus 0.4em\relax International Society for Optics and Photonics, 2004, pp.
  200--219.

\bibitem{uzkent2016integrating}
B.~Uzkent, M.~J. Hoffman, and A.~Vodacek, ``{Integrating Hyperspectral
  Likelihoods in a Multidimensional Assignment Algorithm for Aerial Vehicle
  Tracking},'' \emph{IEEE Journal of Selected Topics in Applied Earth
  Observations and Remote Sensing}, vol.~9, no.~9, pp. 4325--4333, 2016.

\bibitem{han2017efficient}
S.~Han, A.~Fafard, J.~Kerekes, M.~Gartley, E.~Ientilucci, A.~Savakis, C.~Law,
  J.~Parhan, M.~Turek, K.~Fieldhouse \emph{et~al.}, ``Efficient generation of
  image chips for training deep learning algorithms,'' in \emph{Automatic
  Target Recognition XXVII}, vol. 10202.\hskip 1em plus 0.5em minus 0.4em\relax
  International Society for Optics and Photonics, 2017, p. 1020203.

\bibitem{han2017overview}
S.~Han and J.~P. Kerekes, ``Overview of passive optical multispectral and
  hyperspectral image simulation techniques,'' \emph{IEEE Journal of Selected
  Topics in Applied Earth Observations and Remote Sensing}, 2017.

\bibitem{uzkent2015spectral}
B.~Uzkent, M.~J. Hoffman, and A.~Vodacek, ``Spectral validation of measurements
  in a vehicle tracking dddas,'' \emph{Procedia Computer Science}, vol.~51, pp.
  2493--2502, 2015.

\bibitem{uzkent2016real_2}
B.~Uzkent, \emph{Real-time Aerial Vehicle Detection and Tracking using a
  Multi-modal Optical Sensor}.\hskip 1em plus 0.5em minus 0.4em\relax Rochester
  Institute of Technology, 2016.

\bibitem{henriques2015high}
J.~F. Henriques, R.~Caseiro, P.~Martins, and J.~Batista, ``High-speed tracking
  with kernelized correlation filters,'' \emph{IEEE Transactions on Pattern
  Analysis and Machine Intelligence}, vol.~37, no.~3, pp. 583--596, 2015.

\bibitem{hare2016struck}
S.~Hare, S.~Golodetz, A.~Saffari, V.~Vineet, M.-M. Cheng, S.~L. Hicks, and
  P.~H. Torr, ``Struck: Structured output tracking with kernels,'' \emph{IEEE
  transactions on pattern analysis and machine intelligence}, vol.~38, no.~10,
  pp. 2096--2109, 2016.

\bibitem{kalal2012tracking}
Z.~Kalal, K.~Mikolajczyk, and J.~Matas, ``Tracking-learning-detection,''
  \emph{IEEE transactions on pattern analysis and machine intelligence},
  vol.~34, no.~7, pp. 1409--1422, 2012.

\bibitem{lalonde2017fully}
R.~LaLonde, D.~Zhang, and M.~Shah, ``Fully convolutional deep neural networks
  for persistent multi-frame multi-object detection in wide area aerial
  videos,'' \emph{arXiv preprint arXiv:1704.02694}, 2017.

\bibitem{yi2016vehicle}
M.~Yi, F.~Yang, E.~Blasch, C.~Sheaff, K.~Liu, G.~Chen, and H.~Ling, ``Vehicle
  classification in wami imagery using deep network,'' in \emph{SPIE Defense+
  Security}.\hskip 1em plus 0.5em minus 0.4em\relax International Society for
  Optics and Photonics, 2016, pp. 98\,380E--98\,380E.

\bibitem{dalal2005histograms}
N.~Dalal and B.~Triggs, ``Histograms of oriented gradients for human
  detection,'' in \emph{Computer Vision and Pattern Recognition, 2005. CVPR
  2005. IEEE Computer Society Conference on}, vol.~1.\hskip 1em plus 0.5em
  minus 0.4em\relax IEEE, 2005, pp. 886--893.

\bibitem{li2014scale}
Y.~Li and J.~Zhu, ``A scale adaptive kernel correlation filter tracker with
  feature integration,'' in \emph{European Conference on Computer
  Vision}.\hskip 1em plus 0.5em minus 0.4em\relax Springer, 2014, pp. 254--265.

\bibitem{felzenszwalb2010object}
P.~F. Felzenszwalb, R.~B. Girshick, D.~McAllester, and D.~Ramanan, ``Object
  detection with discriminatively trained part-based models,'' \emph{IEEE
  transactions on pattern analysis and machine intelligence}, vol.~32, no.~9,
  pp. 1627--1645, 2010.

\bibitem{bolme2010visual}
D.~S. Bolme, J.~R. Beveridge, B.~A. Draper, and Y.~M. Lui, ``Visual object
  tracking using adaptive correlation filters,'' in \emph{Computer Vision and
  Pattern Recognition (CVPR), 2010 IEEE Conference on}.\hskip 1em plus 0.5em
  minus 0.4em\relax IEEE, 2010, pp. 2544--2550.

\bibitem{galoogahi2013multi}
H.~K. Galoogahi, T.~Sim, and S.~Lucey, ``Multi-channel correlation filters,''
  in \emph{Proceedings of International Conference on Computer Vision}, 2013.

\bibitem{henriques2012exploiting}
J.~F. Henriques, R.~Caseiro, P.~Martins, and J.~Batista, ``Exploiting the
  circulant structure of tracking-by-detection with kernels,'' in
  \emph{Proceedings on European Conference on Computer Vision}, 2012, pp.
  702--715.

\bibitem{krizhevsky2012imagenet}
A.~Krizhevsky, I.~Sutskever, and G.~E. Hinton, ``Imagenet classification with
  deep convolutional neural networks,'' in \emph{Advances in neural information
  processing systems}, 2012, pp. 1097--1105.

\bibitem{simonyan2014very}
K.~Simonyan and A.~Zisserman, ``Very deep convolutional networks for
  large-scale image recognition,'' \emph{arXiv preprint arXiv:1409.1556}, 2014.

\bibitem{danelljan2015convolutional}
M.~Danelljan, G.~Hager, F.~Shahbaz~Khan, and M.~Felsberg, ``Convolutional
  features for correlation filter based visual tracking,'' in \emph{Proceedings
  of the IEEE International Conference on Computer Vision Workshops}, 2015, pp.
  58--66.

\bibitem{held2016learning}
D.~Held, S.~Thrun, and S.~Savarese, ``Learning to track at 100 fps with deep
  regression networks,'' in \emph{European Conference on Computer
  Vision}.\hskip 1em plus 0.5em minus 0.4em\relax Springer, 2016, pp. 749--765.

\bibitem{bertinetto2016fully}
L.~Bertinetto, J.~Valmadre, J.~F. Henriques, A.~Vedaldi, and P.~H. Torr,
  ``Fully-convolutional siamese networks for object tracking,'' in
  \emph{European Conference on Computer Vision}.\hskip 1em plus 0.5em minus
  0.4em\relax Springer, 2016, pp. 850--865.

\bibitem{leal2016learning}
L.~Leal-Taix{\'e}, C.~Canton-Ferrer, and K.~Schindler, ``Learning by tracking:
  Siamese cnn for robust target association,'' in \emph{Proceedings of the IEEE
  Conference on Computer Vision and Pattern Recognition Workshops}, 2016, pp.
  33--40.

\bibitem{mueller2016uav123}
M.~Mueller, N.~Smith, and B.~Ghanem, ``A benchmark and simulator for uav
  tracking,'' in \emph{Proceedings of European Conference on Computer Vision},
  2016, pp. 445--461.

\bibitem{ientilucci2003advances}
E.~J. Ientilucci and S.~D. Brown, ``Advances in wide-area hyperspectral image
  simulation,'' in \emph{AeroSense 2003}.\hskip 1em plus 0.5em minus
  0.4em\relax International Society for Optics and Photonics, 2003, pp.
  110--121.

\bibitem{lowe2004distinctive}
D.~G. Lowe, ``Distinctive image features from scale-invariant keypoints,''
  \emph{International journal of computer vision}, vol.~60, no.~2, pp. 91--110,
  2004.

\bibitem{fischler1987random}
M.~A. Fischler and R.~C. Bolles, ``Random sample consensus: a paradigm for
  model fitting with applications to image analysis and automated
  cartography,'' in \emph{Readings in computer vision}.\hskip 1em plus 0.5em
  minus 0.4em\relax Elsevier, 1987, pp. 726--740.

\bibitem{tang2015multi}
M.~Tang and J.~Feng, ``Multi-kernel correlation filter for visual tracking,''
  in \emph{Proceedings of International Conference on Computer Vision}, 2015,
  pp. 3038--3046.

\bibitem{ma2015long}
C.~Ma, X.~Yang, C.~Zhang, and M.-H. Yang, ``Long-term correlation tracking,''
  in \emph{Proceedings of the IEEE Conference on Computer Vision and Pattern
  Recognition}, 2015, pp. 5388--5396.

\bibitem{bibi2015multi}
A.~Bibi and B.~Ghanem, ``Multi-template scale-adaptive kernelized correlation
  filters,'' in \emph{Proceedings of the IEEE International Conference on
  Computer Vision Workshops}, 2015, pp. 50--57.

\bibitem{rifkin2003regularized}
R.~Rifkin, G.~Yeo, T.~Poggio \emph{et~al.}, ``Regularized least-squares
  classification,'' \emph{Nato Science Series Sub Series {III} Computer and
  Systems Sciences}, vol. 190, pp. 131--154, 2003.

\bibitem{girshick2015fast}
R.~Girshick, ``Fast r-cnn,'' in \emph{Proceedings of the IEEE international
  conference on computer vision}, 2015, pp. 1440--1448.

\bibitem{ren2015faster}
S.~Ren, K.~He, R.~Girshick, and J.~Sun, ``Faster r-cnn: Towards real-time
  object detection with region proposal networks,'' in \emph{Advances in neural
  information processing systems}, 2015, pp. 91--99.

\bibitem{dai2016r}
J.~Dai, Y.~Li, K.~He, and J.~Sun, ``R-fcn: Object detection via region-based
  fully convolutional networks,'' in \emph{Advances in neural information
  processing systems}, 2016, pp. 379--387.

\bibitem{yu2015multi}
F.~Yu and V.~Koltun, ``Multi-scale context aggregation by dilated
  convolutions,'' \emph{arXiv preprint arXiv:1511.07122}, 2015.

\bibitem{jia2014caffe}
Y.~Jia, E.~Shelhamer, J.~Donahue, S.~Karayev, J.~Long, R.~Girshick,
  S.~Guadarrama, and T.~Darrell, ``Caffe: Convolutional architecture for fast
  feature embedding,'' \emph{arXiv preprint arXiv:1408.5093}, 2014.

\bibitem{redmon2016yolo9000}
J.~Redmon and A.~Farhadi, ``Yolo9000: better, faster, stronger,'' \emph{arXiv
  preprint arXiv:1612.08242}, 2016.

\bibitem{liu2016ssd}
W.~Liu, D.~Anguelov, D.~Erhan, C.~Szegedy, S.~Reed, C.-Y. Fu, and A.~C. Berg,
  ``Ssd: Single shot multibox detector,'' in \emph{European conference on
  computer vision}.\hskip 1em plus 0.5em minus 0.4em\relax Springer, 2016, pp.
  21--37.

\bibitem{poore1994multidimensional}
A.~B. Poore, ``Multidimensional assignment formulation of data association
  problems arising from multitarget and multisensor tracking,''
  \emph{Computational Optimization and Applications}, vol.~3, no.~1, pp.
  27--57, 1994.

\end{thebibliography}
}
%




\end{document}